\documentclass[lettersize,journal]{IEEEtran}
\usepackage{amsmath,amsfonts}
\usepackage{algorithmic}
\usepackage{algorithm}
\usepackage{array}
\usepackage[caption=false,font=normalsize,labelfont=sf,textfont=sf]{subfig}
\usepackage{textcomp}
\usepackage{stfloats}
\usepackage{url}
\usepackage{verbatim}
\usepackage{graphicx}
\usepackage{cite}
\usepackage{subfiles}
\usepackage{graphicx}
\usepackage{amsmath}
\usepackage{amssymb}
\usepackage{multirow}
\usepackage{makecell}
\usepackage{color}
\usepackage{bm}
\usepackage[pagebackref,breaklinks,colorlinks]{hyperref}
\usepackage{marvosym}
\usepackage{xcolor}
\usepackage{booktabs}
\usepackage{colortbl}
\usepackage{threeparttable}

\usepackage[capitalize]{cleveref}
\crefname{section}{Sec.}{Secs.}
\Crefname{section}{Section}{Sections}
\Crefname{table}{Table}{Tables}
\crefname{table}{Tab.}{Tabs.}

\newcommand{\FIG}[1]{\cref{#1}}
\newcommand{\TABLE}[1]{\cref{#1}}
\newcommand{\EQUA}[1]{\cref{#1}}
\newcommand{\SECTION}[1]{\cref{#1}}

\newcommand{\MODEL}{Unnamed Model}

\newcommand{\ETAL}{{\emph{et al.}}}

\newcommand{\IE}{{\emph{i.e.}}}

\newcommand{\SOCIALLSTM}{Social-LSTM}
\newcommand{\SOCIALLSTMCITE}{\cite{alahi2016social}}

\newcommand{\SIMAUGCITE}{\cite{liang2020simaug}}
\newcommand{\PECNET}{PECNet}
\newcommand{\PECNETCITE}{\cite{mangalam2020not}}

\newcommand{\GARDENCITE}{\cite{liang2020garden}}
\newcommand{\STAR}{STAR}
\newcommand{\STARCITE}{\cite{yu2020spatio}}

\newcommand{\TRAJECTRONPP}{Trajectron++}

\newcommand{\TRAJECTRONPPCITE}{\cite{salzmann2020trajectron}}
\newcommand{\YNET}{$\textsf{Y}$-net}
\newcommand{\YNETCITE}{\cite{mangalam2020s}}

\newcommand{\SPECTGNN}{SpecTGNN}
\newcommand{\SPECTGNNCITE}{\cite{cao2021spectral}}

\newcommand{\AGENTFORMER}{AgentFormer}
\newcommand{\AGENTFORMERCITE}{\cite{yuan2021agentformer}}

\newcommand{\VMODEL}{V$^2$-Net}

\newcommand{\VCITE}{\cite{wong2022view}}

\newcommand{\MUSEVAE}{MUSE-VAE}
\newcommand{\MUSEVAECITE}{\cite{lee2022muse}}

\newcommand{\EVMODEL}{E-V$^2$-Net}

\newcommand{\EVCITE}{\cite{wong2023another}}

\newcommand{\SCMODEL}{SocialCircle}
\newcommand{\SCCITE}{\cite{wong2023socialcircle}}
\newcommand{\MSRL}{MSRL}
\newcommand{\MSRLCITE}{\cite{wu2023multi}}

\newcommand{\LGTRAJ}{LG-Traj}
\newcommand{\LGTRAJCITE}{\cite{chib2024lg}}

\newcommand{\UPDD}{UPDD}
\newcommand{\UPDDCITE}{\cite{liu2024uncertainty}}
\newcommand{\SCPMODEL}{SocialCircle+}
\newcommand{\SCPCITE}{\cite{wong2024socialcircle+}}
\newcommand{\UEN}{UEN}
\newcommand{\UENCITE}{\cite{su2024unified}}
\newcommand{\MSTIP}{MS-TIP}
\newcommand{\MSTIPCITE}{\cite{chib2024ms}}
\newcommand{\SMEMO}{SMEMO}
\newcommand{\SMEMOCITE}{\cite{marchetti2024smemo}}
\newcommand{\DICE}{Dice}
\newcommand{\DICECITE}{\cite{choi2024dice}}
\newcommand{\PPT}{PPT}
\newcommand{\PPTCITE}{\cite{lin2024progressive}}
\newcommand{\AFFLN}{AgentFormer-FLN}
\newcommand{\AFFLNCITE}{\cite{xu2024adapting}}
\newcommand{\SOPERMODEL}{SoperModel}
\newcommand{\SOPERMODELCITE}{\cite{yang2025sopermodel}}
\newcommand{\REMODEL}{\emph{Resonance}}
\newcommand{\RECITE}{\cite{wong2024resonance}}

\newcommand{\REVMODEL}{\emph{Reverberation}}
\newcommand{\REVCITE}{\cite{wong2025reverberation}}

\renewcommand{\MODEL}{\emph{Encore}}

\newcommand{\C}[1]{{\textcolor[HTML]{0000FF}{#1}}}
\newcommand{\CC}[1]{{\underline{\C{#1}}}}
\newcommand{\X}[1]{{\textcolor[HTML]{FF4F4F}{#1}}}
\newcommand{\TODO}[1]{\colorbox{yellow}{TODO}}
\newcommand{\hatbf}[1]{\hat{\mathbf{#1}}}

\newcommand{\EGO}{{\mathrm{ego}}}
\newcommand{\FINAL}{{\mathrm{f{}in}}}
\newcommand{\BIASED}{{\mathrm{biased}}}

\newcommand{\FC}{\mathrm{fc}}
\newcommand{\RELU}{\mathrm{ReLU}}
\newcommand{\TANH}{\tanh}

\newcommand{\NEWHALFLINE}{\vspace{0.5em}\noindent}

\begin{document}

\title{
    Encore: Conditioning Trajectory Forecasting via Biased Ego Rehearsals
}

\author{
    Conghao Wong,
    Ziqian Zou,
    and~Xinge You,~\IEEEmembership{Senior Member,~IEEE}
    \thanks{
        This work was supported in part by the National Natural Science Foundation of China under Grant 62172177.
        \emph{(Corresponding author: Xinge You.)}
    }
    \thanks{
        Conghao Wong, Ziqian Zou, and Xinge You are with Huazhong University of Science and Technology, Wuhan 430074, Hubei, China.
        (Email: \mbox{conghaowong@icloud.com},
        ziqianzoulive@icloud.com,
        youxg@mail.hust.edu.cn).
    }
    \thanks{
        Code is available at \url{https://github.com/cocoon2wong/Encore}.
    }
}

\markboth{Journal of \LaTeX\ Class Files,~Vol.~14, No.~8, August~2021}%
{Shell \MakeLowercase{\textit{et al.}}: A Sample Article Using IEEEtran.cls for IEEE Journals}


\maketitle

\begin{abstract}

Learning and representing the subjectivities of agents has become a challenging but crucial problem in the trajectory prediction task.
Such subjectivities not only present specific spatial or temporal structures, but also are anisotropic for all interaction participants.
Despite great efforts, it remains difficult to explicitly learn and forecast these subjectivities, let alone further modulate models' predictions through a specific ego's subjectivity.
Inspired by prefactual thoughts in psychology and relevant theatrical concepts, we interpret such subjectivities in future trajectories as the continuous process from rehearsal to encore.
In the rehearsal phase, the proposed ego predictor focuses on how each ego agent learns to derive and direct a set of explicitly biased rehearsal trajectories for all participants in the scene from the short-term observations.
Then, these rehearsal trajectories serve as immediate controls to condition final predictions, providing direct yet distinct ego biases for the prediction network to simulate agents' various subjectivities.
Experiments across datasets not only demonstrate a consistent improvement in the performance of the proposed \MODEL~trajectory prediction model but also provide clear interpretability regarding subjectivities as biased ego rehearsals.

\end{abstract}




\section{Introduction}

\begin{figure}[tbp]
    \centering
    \includegraphics[width=1.0\linewidth]{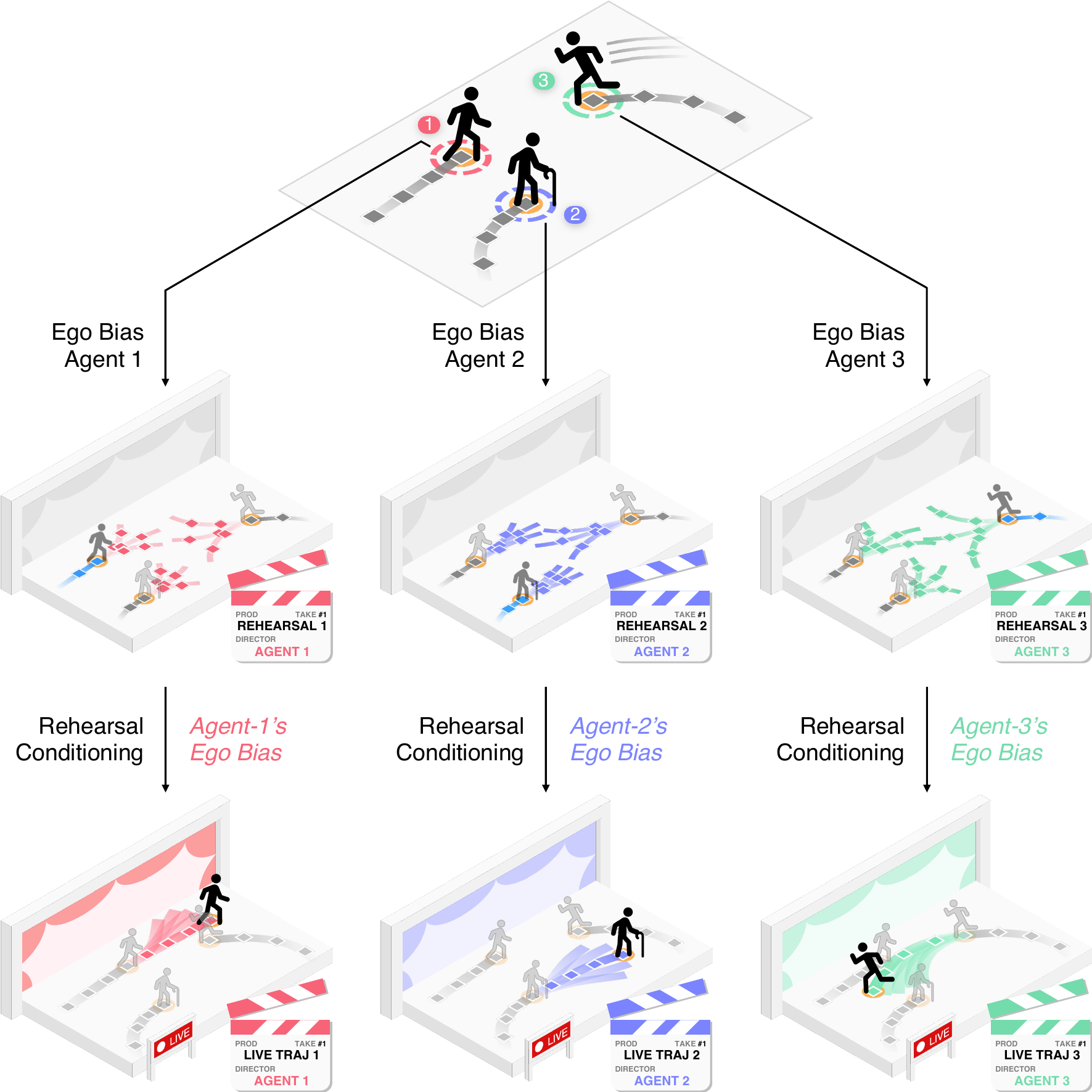}
    \caption{
        Motivation illustration.
        The processes of both social interactions and trajectory planning are mostly \emph{structured} and \emph{anisotropic}, starting from how agents uniquely perceive and then construct their own biased interactive contexts.
        Inspired by theatrical concepts, we restructure trajectory prediction into two consecutive phases.
        Each agent first uses its own ego bias to consider all participants in the scene, directing a set of short but structured \emph{biased rehearsal} trajectories as its exclusive interaction context.
        Then, it produces the final live performance based on these unique rehearsals, thereby conditioning interactions and trajectories on specific ego biases.
    }
    \label{fig_intro}
\end{figure}

\IEEEPARstart{T}{rajectory} prediction is a computer vision task that forecasts agents' possible future movements by considering their historical status and potential contexts of social interactions \SOCIALLSTMCITE.
Understanding and predicting human intentions and social tendencies is not only a crucial foundation for sociological research, but also essential for various advanced downstream applications, such as intelligent transportation systems \cite{chen2025dstigcn,zhou2024trajpred,kothari2023safety}, interactive entertainment \cite{rempe2023trace,yue2024human,bae2025continuous}, and virtual reality or augmented reality \cite{yagi2018future,patel2025uniegomotion}.

Despite great efforts and numerous attempts by the community, the modeling of diverse social interactions as well as uncertain future choices of different agents remains challenging for this task, especially due to the \emph{inherent subjectivity} manifested by the agents to be forecasted.
Unlike other deterministic computer vision tasks, this means that the forecasting of trajectories requires not only the understanding and the encoding of objective trajectory representations, but also these \emph{structured}, \emph{anisotropic} subjectivities.
Here, structured indicates that these subjectivities are mostly associated with specific interaction contexts rather than randomly distributed, whether temporally or spatially.
Meanwhile, anisotropic means that preferences regarding trajectories and interactions are unique to each agent and asymmetrical among interaction participants, rather than being uniformly and universally shared.

Current researchers have attempted to simulate such subjectivities in two main ways, the direct learning of social structures and the advanced sampling of noise vectors.
On one hand, some researchers try to introduce learnable directed or undirected graphs \cite{li2025unified,chen2025dstigcn,li2022graph} to encode agents' subjectivities as specific structured social roles for interacting with others and planning trajectories.
Several group-based methods \cite{xu2022groupnet,bae2022learning,zou2024who} have also been proposed to further characterize such structured roles hierarchically.
However, in most of these approaches, the learned social structures are usually shared by all the agents in the prediction scene simultaneously, which brings further difficulties in distinguishing the anisotropic subjectivities presented by different agents.

On the other hand, with the rapid development of generative neural networks, some other researchers have attempted to interpret such subjectivities as the results of randomness within agents' behaviors.
From earlier GAN-based \cite{rossi2021human,gupta2018social} or VAE-based \cite{lee2022muse,xu2022socialvae} methods to the recent diffusion-model-based ones \cite{bae2024singulartrajectory,li2024bcdiff}, these researchers mainly consider the planning of trajectories as the sampling of features on the specifically constructed or learnable behavioral manifolds, therefore attempting to cover as many trajectories as possible to approximate agents' subjectivities.
Although these methods may have the ability to ensure the anisotropy of samplings (such as through specific sampling strategies \YNETCITE), most of them still lack structural prior considerations of agents' subjectivities when sampling from the constructed feature space, making them difficult to apply to different prediction scenarios while imposing fairly strict requirements on designing specific sampling methods.

Directly combining these two kinds of approaches not only significantly increases network complexity but also makes it difficult to preserve the advantages from both sides.
To address the above requirements, interpreting agents' subjectivities through explicitly controllable yet inherently anisotropic structures has become the main consideration of this manuscript.
Such structures should not only be able to distinguish these diverse anisotropic subjectivities, but also fully preserve agents' social roles and their corresponding social contexts.
It is challenging to directly design such mathematical structures.
However, as trajectory forecasting is fundamentally a human-centric task, leveraging certain human behavioral intuitions and established workflows from other fields may further provide novel insights, like the previous echolocation-inspired interaction-modeling approaches \cite{wong2023socialcircle,wong2024socialcircle+}.

Interestingly, the above anisotropic yet structured requirements are indirectly manifested not only in our human intuitions as pedestrians but further in current industrialized theatrical workflows.
For example, when pedestrians plan their future trajectories in a relatively complex scenario, they may first generate rough but quick snapshots to help rapidly figure out ongoing dynamics as well as states of every interaction participant in the immediate future.
In psychology, this type of snapshot is referred to as \emph{mental simulations} or the more specific \emph{prefactual thoughts} \cite{epstude2016prefactual}.
Such simulations rely heavily on personal experiences.
Although they may not accurately capture and locate every detail, such simulations still mostly preserve the overall structures of the current interactions while providing highly \emph{biased} imaginations of the interactive contexts in the near future, thereby providing direct priors for planning pedestrians' subsequent motions.
In this process, each agent's unique subjectivity is reflected in these imaginations, which contain not only complete interaction structures but also distinct asymmetrical and anisotropic cognitive differences from pedestrians' individual experiences.
Similarly, this process has even been formalized into distinct steps in the theatrical industry.
When finalizing a script, each writer may direct their own rehearsal takes based on their unique perspective, even when all writers are working within the same scene context.
These rehearsals are asymmetrical, and could reflect strong subjective preferences of different writers.
Subsequently, these explicit rehearsals serve as strong prior conditions for the final production.
Also, the final script is rarely a direct adoption of any single rehearsal, but rather a concatenation of the most compelling segments from each one.

Here, both mental simulations and rehearsal processes provide direct structural insights to simulate and quantify unique subjectivities presented by the agents in the trajectory prediction task.
As illustrated in \FIG{fig_intro}, we borrow the concept of \emph{rehearsal} to directly and explicitly learn subjectivities with specific structures and anisotropic properties, by simulating rehearsal scenarios and directing the corresponding rehearsal trajectories with unique \emph{ego biases} for each agent, and finally using such rehearsals as conditions to modulate the final forecasted trajectories.
Specifically, different from features either aggregated from graph structures or sampled from high-level feature spaces, such rehearsal trajectories serve as explicit contexts for representing agents' subjectivities, manifested as a set of short-term forecasted trajectories for each agent in the scene, directed by each specific ego agent.

In detail, as shown in \FIG{fig_intro}, we restructure trajectory prediction
into two consecutive phases.
Before the final forecasting, corresponding to the above mental simulations, we first trim historical observations to a shorter temporal horizon.
Upon such snapshot observations, we then design an ego predictor that explicitly learns and models ego biases while additionally accounting for the directions of such asymmetric biases (\IE, we regard that bias $B^{j \leftarrow i}$ sourced from a specific ego $i$ towards a neighbor $j$ is not strictly the same as the flipped $B^{i \leftarrow j}$).
The ego predictor predicts all potential multimodal short-term trajectories for all agents in the scene, simulating the rehearsal process for considering each ego-agent's future interaction selections and trajectory plans under specific subjective biases.
This means that rehearsal trajectories not only fully preserve the interactive structures (manifested as a set of trajectories with spatial relationships) but also reflect the anisotropy of agents' subjectivities, thereby serving as a solid basis for directly controlling how the final predictor interprets agents' subjectivities.
In particular, unlike current waypoint-based approaches, rehearsal trajectories do not require the selection of any single rehearsal or strict correspondences with the final prediction.
Rather, they are allowed to be exploratory or even imperfect, and are treated as an explicit prior that conditions the final prediction from both the feature and trajectory levels, incorporating the unique aspects while discarding the ``plain'' segments across all rehearsal trajectories.

From the perspective of the ego agents themselves, this process can be understood as their first directing and overseeing the entire rehearsal process, then analyzing and selecting the rehearsals to establish the final conditions corresponding to their subjectivities, and finally performing the rehearsed social behaviors and trajectory planning.
Therefore, we refer to such a proposed two-phase trajectory prediction model as \MODEL, indicating the \emph{additional performance}.

In summary, we contribute
(1) The ego predictor that learns and forecasts short-term rehearsal trajectories with specific ego-agents' biases for considering interactions and future plans, explicitly representing their structured and anisotropic subjectivities.
(2) The \MODEL~trajectory prediction model integrated with the ego predictor and the corresponding ego loss to forecast trajectories under specific ego biases, conditioned by the forecasted biased rehearsals from the immediate trajectory and further feature selection levels.
(3) Experiments and discussions on multiple datasets that not only validate the consistent performance improvements, but also provide further interpretations of how agents' subjectivities are distributed and finally condition their future selections.

\section{Related Works}

\subsection{Trajectory Prediction and Social Interactions}

Trajectory prediction can be treated as one of the sequential forecasting tasks.
With the development of sequential neural networks like RNNs and LSTMs, researchers have attempted to bring trainable fashions into this task to replace the manual-rule based methods like Social Force \cite{helbing1995social}.
\SOCIALLSTM~\SOCIALLSTMCITE~first introduce LSTM to this task, using the shared trainable average-pooled feature to represent interaction contexts.
Subsequently, such sequential-feature-based shared social representations \cite{zhang2020social,huang2021lstm,zhang2019sr,bisagno2018group,xue2018ss} became the first choice for most researchers, as they could provide social contexts without the requirement of further manual designs.

Until the advent of graph networks \cite{kipf2016semi}, researchers recognized a significant weakness of the shared pooling-based approaches, where they mostly fail to account for the detailed spatial structures and positional relationships between interaction participants.
Since then, graph convolution networks \cite{chen2025dstigcn,shi2021sgcn}, graph attention networks \cite{lv2023skgacn,liu2021avgcn,kosaraju2019social}, and further spatial-temporal graphs \cite{mohamed2020social,huang2019stgat} or even spatial-frequential graphs \cite{cao2021spectral,cao2020spectral} have become current main choices for representing interaction contexts.
For most of these graph-based methods, they use one agent's temporal feature as the node, and employ their uniquely designed similarity metrics and features as edges to construct either directed or undirected graph structures that describe the interaction contexts.

Recent methods have moved away from these temporal features, instead leveraging additional knowledge from large language models (LLMs) to treat trajectory prediction as a natural language-based question-answering (QA) task \cite{chib2024lg,bae2025social}.
They describe the scene structure in natural language, specify the motion state of each agent (coordinates embedded via specific templates), and use this additional language knowledge to infer potential interactions between agents, achieving progress on some zero-shot scenarios.

However, whether using graphs or LLMs, although these methods can effectively preserve the structures of interactions, they rarely account for the asymmetric or anisotropic properties of social interactions.
For example, for a given prediction scenario, researchers mainly construct a single shared graph structure, meaning that interaction representations remain shared regardless of which agent is being predicted as the ego agent or any unconcerned neighbors.
Similarly, even for LLM-based methods, their symmetric state templates can hardly represent anisotropic and subjectively biased subjectivities presented by agents to be forecasted.
Therefore, finding a structure that has the ability to demonstrate the anisotropy and biased interactive contexts has become one of our main concerns.

\subsection{Stochastic Trajectory Prediction and Subjectivities}

In addition to the above direct modeling of social interactions, generating multimodal possible trajectories for all agents has become another main focus of the trajectory prediction task.
These approaches are also referred to as \emph{stochastic methods} compared to early works that could only forecast one deterministic trajectory, like the \SOCIALLSTM~that predicts one trajectory for each agent as their statistical average behaviors.

Earlier researchers mainly use generative neural networks to forecast multiple socially acceptable trajectories through feature samplings.
GAN-based models \cite{kothari2023safety,sadeghian2019sophie,gupta2018social} introduce adversarial competitions between the generator (feature encoder) and the discriminator, thus learning a trainable feature space to consider all agents' motion modes or patterns.
Such methods require carefully designed optimization strategies and loss functions, such as the variety loss \cite{kothari2023safety} to constrain the feature space.
As alternatives, VAE-based methods \cite{chib2024ms,lee2022muse,xu2022socialvae} have relatively low requirements for such intermediate features, which are required to satisfy the standardized Gaussian distribution.
These methods achieve multimodal generations by learning to encode and decode observations and predictions in this specific feature space.
Although some methods have proposed new sampling measures or waypoint control conditions \YNETCITE, it is still challenging for either GAN or VAE-based methods to explicitly preserve the structured interactive context within the network, due to the unavoidable information compressions during the feature sampling process.
Recent diffusion-based models \cite{bae2024singulartrajectory,li2024bcdiff,mao2023leapfrog} place greater focus on learning the evolutionary relationships between different samples.
By learning to encode noise vectors in multiple continuous steps, they produce multiple predictions by sampling features between existing samples in an interpolative manner, thereby overcoming the inherent limitations of discrete trajectory data.

While current stochastic trajectory prediction methods do have the ability to achieve the multi-generation goal, it is still difficult for them to account for the structured interactive context during samplings.
Even for recent diffusion models, while the iterative process can simulate the evolution of features (noise) from one sample to another, real-world interactions are often unique and may not satisfy the locally linearly interpolable property.
This means that preserving structured and anisotropic agent subjectivities when generating trajectories remains challenging for these stochastic models, since such samplings may disrupt the structure of the interaction contexts, whether temporally or spatially, thereby limiting the capabilities for fully simulating agents' unique subjectivities.

\NEWHALFLINE
Therefore, to best simulate agents' subjectivities, it is necessary not only to describe and encode interactions as anisotropic representations, but also to maintain the unique yet structured interaction context of each agent beyond the unstructured noise sampling.
Inspired by psychology and theatrical concepts, we use explicit, asymmetric rehearsal trajectories with specific ego biases as the direct intermediate variables to condition predictions through joint optimization.
Also, such rehearsals are not required to be scored or ranked.
Rather, they are integrated and considered jointly as structured interaction conditions.
In other words, the final predictions do not need to strictly correspond to or be limited by these rehearsals, which is the key distinction compared to other methods.

\section{Method}

\begin{figure*}[tbp]
    \centering
    \includegraphics[width=1.0\linewidth]{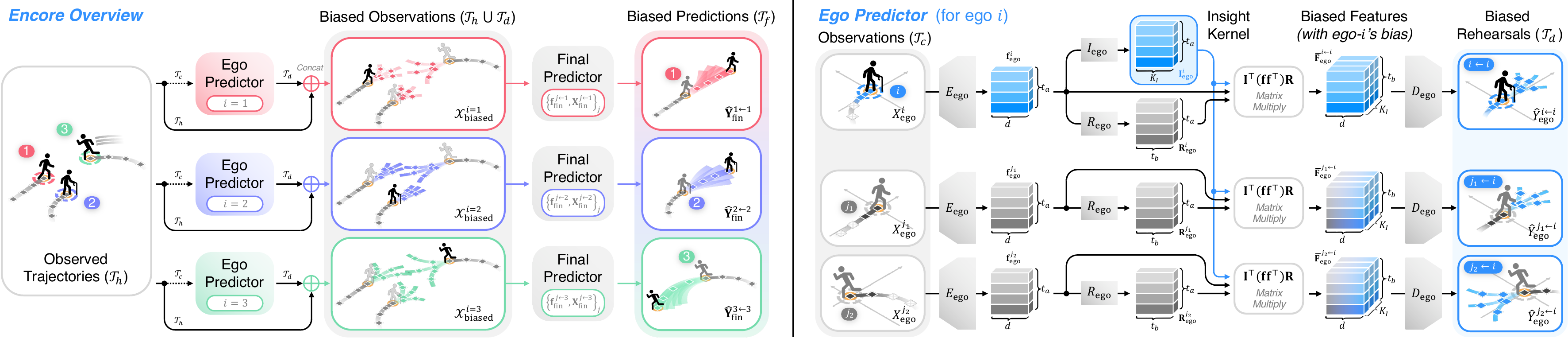}
    \caption{
        Overall computation pipeline of the proposed \MODEL~model (left) and detailed structures of the ego predictor (right) when running the regular forecasting.
        For the ego predictor, an additional computation will be performed on observation and prediction periods $\mathcal{T}_a$ and $\mathcal{T}_b$ (corresponding to periods $\mathcal{T}_c$ and $\mathcal{T}_d$ in the figure) to learn ego biases under the joint optimization of the proposed ego loss (\EQUA{eq_ego_loss}) and the regular $\ell_2$ loss.
    }
    \label{fig_method_overall}
\end{figure*}

The proposed \MODEL~trajectory prediction model consists of two main parts, the \emph{Ego Predictor} and the \emph{Final Predictor}.
Both predictors collaborate to achieve conditioned predictions, guided by the rehearsal trajectories (\IE, biased short-term future predicted trajectories) \emph{directed} by each ego agent.
As illustrated in \FIG{fig_method_overall,fig_method_horizons}, before the final forecasting, the ego predictor first provides a set of biased short-term rehearsals for the ego agent itself and all its neighbors, learning and forecasting how each ego agent uniquely considers its future trajectory plans as well as interaction contexts.
These rehearsals will be concatenated with their original observations and then fed to the final predictor with specific conditioning rules, achieving the goal of asymmetric bias-conditioned trajectory prediction.
This section first formulates the proposed ego predictor, then introduces how the \MODEL~model is constructed and trained.

\subsection{Formulations}
\label{sec_formulations}

Denote the 2D position of any agent $i$ at some time $t$ as $\mathbf{p}^i_t \in \mathbb{R}^2$.
We consider a set of $t_h$ equal-interval-sampled (denote this interval as $\Delta t$) historical observation steps $\mathcal{T}_h = \{1, 2, ..., t_h\}$ and any ego agent $i$ (the agent to be forecasted, from the set of all agents $\mathcal{N} = \{1, 2, ..., N_a\}$ in the prediction scene) when forecasting.
Given a set of the following $t_f$ future prediction steps $\mathcal{T}_f = \{t_h+1, t_h+2, ..., t_h+t_f\}$, our final trajectory prediction goal (\IE, the goal of the final predictor) is to forecast possible future trajectories $\hatbf{Y}^i_f = (\hatbf{p}^i_{t_h + 1}, \hatbf{p}^i_{t_h + 2}, ..., \hatbf{p}^i_{t_h + t_f})^\top \in \mathbb{R}^{t_f \times 2}$, according to all agents' observed trajectories $\mathcal{X}_h = \{\mathbf{X}^j_h | j \in \mathcal{N}\}$, where each $\mathbf{X}^j_h = (\mathbf{p}^j_1, \mathbf{p}^j_2, ..., \mathbf{p}^j_{t_h})^\top \in \mathbb{R}^{t_h \times 2}$.

\begin{table}
    \caption{
        Symbols and network structures used in the \MODEL~model
    }
    \label{tab_symbols}
    \centering
    \begin{tabular}{cllc}
        \toprule
        Symbols & Layers (Using $[\cdot, \cdot]$ to denote concatenate) \\

        \midrule
        $E_\EGO$ & $\left[X^i_\EGO, X^j_\EGO\right] \to \FC(d, \RELU) \to \FC(d, \TANH) $ \\
         & $\quad \to \mbox{Transformer}(d, \mbox{2 heads})\mbox{*2} \to \left[\mathbf{f}^i_\EGO, \mathbf{f}^j_\EGO\right]$ \\[1.1ex]
        $R_\EGO$ & $\mathbf{f}^j_\EGO \to \FC(d, \RELU)\mbox{*2} \to \FC(t_b, \TANH) \to \mathbf{R}^j_\EGO$ \\[1.1ex]
        $I_\EGO$ & $\mathbf{f}^i_\EGO \to \FC(d, \RELU)\mbox{*2} \to \FC(K_I, \TANH) \to \mathbf{I}^i_\EGO$ \\[1.1ex]
        $D_\EGO$ & $\overline{\mathbf{F}}^{j \leftarrow i}_\EGO \to \FC(2) \to \hat{Y}^{j \leftarrow i}_\EGO$ \\

        \bottomrule
    \end{tabular}

\end{table}

\subsection{Biased Ego Predictor}

Unlike the final predictor, the key idea of the proposed biased ego predictor (or simply referred to as \emph{Ego Predictor}) is to provide diverse biased short-term future predictions for all neighboring agents $j \in \mathcal{N}$ from some ego agent $i$'s point of view.
Its core consideration is that the same neighbor's trajectory or interactive behaviors could be interpreted differently from different egos' points of view in actual prediction scenes.
Here, such point of view is not equivalent to the mathematical coordinate transform to handle all neighbors in the ego-centric space, but provides agent-specific \emph{biases} for considering its neighbors' possible future movements or interactions, \IE, biased future insights.
In short, this ego predictor tries to learn how each ego agent uniquely considers its neighbors, under the assumption of ``\emph{Suppose I were this ego, how would I perceive and consider my interactions with these neighbors?}''.
Correspondingly, the main goal of the proposed ego predictor is to learn such asymmetric ego biases and finally to predict such biased ``rehearsal'' trajectories.
This section first distinguishes the prediction horizons of ego predictor and final predictor, then introduces this asymmetric structure to learn and to forecast distinct ego biases in detail.

\NEWHALFLINE\subsubsection{Prediction Horizons}
\label{sec_method_horizon}

\begin{figure}[tbp]
    \centering
    \includegraphics[width=1.0\linewidth]{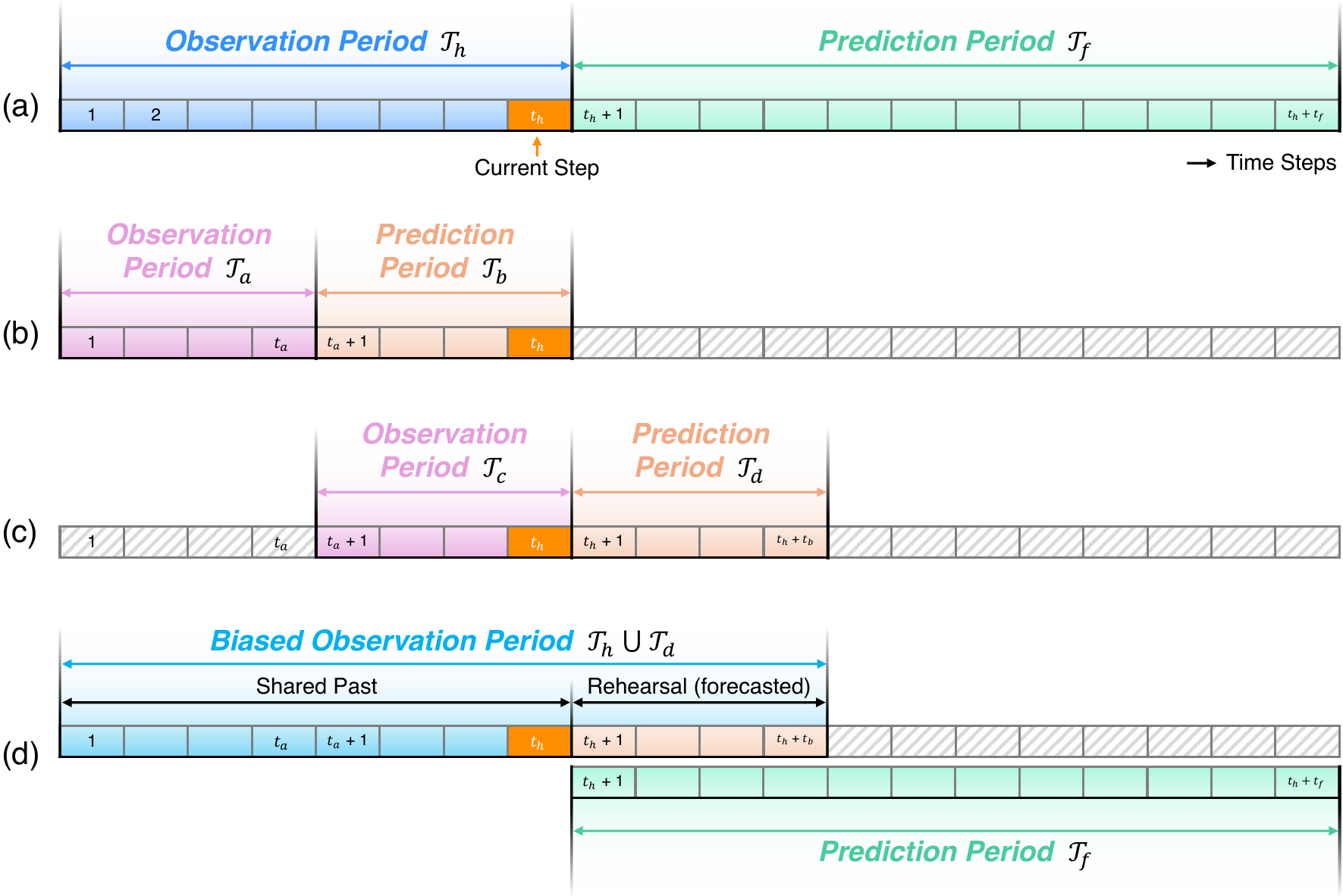}
    \caption{
        Illustration of different prediction horizons used in the proposed \MODEL~model.
        (a) shows the regular periods $\{\mathcal{T}_h, \mathcal{T}_f\}$.
        (b) and (c) indicate the periods $\{\mathcal{T}_a, \mathcal{T}_b\}$ and $\{\mathcal{T}_c, \mathcal{T}_d\}$ used during the training and actual inference phases of the ego predictor, respectively.
        (d) visualizes the final biased observation and prediction periods used in the final predictor.
    }
    \label{fig_method_horizons}
\end{figure}

The ego predictor is a lightweight trajectory prediction model.
It does not aim at maximizing trajectory accuracy, but at learning and forecasting possible biases presented by different ego agents for considering their own \emph{anisotropic} interaction context.
To learn such biases, as illustrated in \FIG{fig_method_horizons}, prediction horizons of the ego predictor are set differently from the standard $\{\mathcal{T}_h, \mathcal{T}_f\}$ in regular prediction models.
Agents cannot access how others will actually behave in the future in real-world scenes.
Instead, for any ego agent, its ego bias and interaction decisions are exclusively displayed and formed from the shared past with its neighbors until the current step $t = t_h$.
Thus, during the bias-learning process, as shown in \FIG{fig_method_horizons} (b), observation and prediction periods $\{\mathcal{T}_a, \mathcal{T}_b\}$ of the ego predictor are set to
\begin{align}
    \mathcal{T}_a &= \{1, ..., t_a\}, \\
    \mathcal{T}_b &= \{t_a + 1, ..., t_h\}.
\end{align}
Here, these bias-learning (training) periods satisfy
\begin{equation}
    \left\{\begin{aligned}
        & \mathcal{T}_a \cup \mathcal{T}_b = \mathcal{T}_h, \\
        & \min \mathcal{T}_b - \max \mathcal{T}_a = 1.
    \end{aligned}\right.
\end{equation}
In other words, in the training phase, the ego predictor is set to predict $t_b = \| \mathcal{T}_b \| = t_h - t_a$ steps of short-term future trajectories according to $t_a = \| \mathcal{T}_a \|$ short observations, thus making full use of all $t_h$ observations for learning potential ego biases, while avoiding the requirement of groundtruth trajectories during the final prediction period $\mathcal{T}_f$.

Accordingly, as illustrated in \FIG{fig_method_horizons} (c), after training, the ego predictor works according to the last $t_a$ steps in the entire observation period $\mathcal{T}_h$ and makes predictions in the following $t_b$ steps.
Compared to the former training phase, its observation and prediction periods have been simultaneously forward-shifted to align with the current observation step $t = t_h$.
Although differences between the training and actual inference phases might introduce potential data shift risks, the overlap of periods $\mathcal{T}_d$ and $\mathcal{T}_f$ (visualized in \FIG{fig_method_horizons} (d)) bridges the ego predictor to the final predictor during the joint optimization (See explanations in \EQUA{eq_rehearsal_concat} and \SECTION{sec_method_conditioning}).
For a clear representation, this observation period $\mathcal{T}_c$ and the corresponding prediction period $\mathcal{T}_d$ are denoted as
\begin{align}
    \label{eq_ego_periods}
    \mathcal{T}_c &= \left\{t_h - t_a + 1, ..., t_h\right\}, \\
    \mathcal{T}_d &= \left\{t_h+1, ..., t_h+t_b\right\}.
\end{align}
Simply, we have $\| \mathcal{T}_c \| = t_a$ and $\| \mathcal{T}_d \| = t_b$.
In addition, the (final) prediction period $\mathcal{T}_f$ is usually not shorter than the observation period $\mathcal{T}_h$ under common trajectory prediction settings ($t_h \leq t_f$), which means that we have $\mathcal{T}_d \subset \mathcal{T}_f$.
Thus, we refer to trajectories forecasted by the ego predictor over period $\mathcal{T}_d$ as the \emph{short-term future rehearsals}, short for \emph{rehearsals}.
These rehearsals are the main conditions for the final predictor to take into account different agents' ego biases.
We will introduce how these components are designed one by one in the following sections.

\NEWHALFLINE\subsubsection{Biased Ego Rehearsals}

We first introduce the core operation to make the ego predictor available to forecast any neighboring agent-$j$'s trajectories corresponding to any different ego-agent-$i$'s bias ($i, j \in \mathcal{N}$).
For such ego-neighbor pair $(i, j)$, due to the differences between the training and actual inference phases, we use variables $X^i_\EGO, X^j_\EGO \in \mathbb{R}^{t_a \times 2}$ to denote input trajectories of the ego predictor, and use variable $\hat{Y}^{j \leftarrow i}_\EGO$ to denote its biased prediction.
For example, we have $X^i_\EGO = \mathbf{X}^i_a, X^j_\EGO = \mathbf{X}^j_a$ only when training, while $X^i_\EGO = \mathbf{X}^i_c, X^j_\EGO = \mathbf{X}^j_c$ otherwise\footnote{
    Subscripts of trajectory symbols denote their periods.
    For example, $\mathbf{X}^i_a$ denotes the observed trajectory during period $\mathcal{T}_a$, and it has $\|\mathcal{T}_a\| = t_a$ steps.
}.

For ego $i$ and neighbor $j$, the ego predictor forecasts neighbor-$j$'s trajectory $\hat{Y}^{j \leftarrow i}_\EGO$ according to both ego-$i$'s and neighbor-$j$'s observations $X^i_\EGO, X^j_\EGO$.
Here, we only consider the direct social interactions between $i$ and $j$, \IE, the first-order social interactions \cite{kim2024higher}.
This helps to simulate the direct and rapid interaction rehearsals by humans in real life, omitting all other indirect or minor interaction events while leaving the full social interactions to be handled by the final predictor.
Before forecasting, the linear part of both $X^i_\EGO$ and $X^j_\EGO$ will be separated, making the learned ego biases able to focus more on the non-linear trajectory dynamics \RECITE.
All computations within the ego predictor are then conducted on the ``clean'' non-linear part of these trajectories, by translating all trajectories so that the origin of all coordinates ($\mathbf{p}^i_t$ or $\mathbf{p}^j_t$) lies at the last observation position of the ego $i$, thus better learning and forecasting how the interaction bias of ego $i$ has been or will be presented.
Note that to reduce complexities of intermediate symbols and notations, equations for the separation and restoration of linear parts (linear fit in $X^i_\EGO, X^j_\EGO$ and linear prediction in $\hat{Y}^j_\EGO$, using linear least squares) and translations are omitted for conciseness in this manuscript (see details in the previous work \RECITE).

Then, trajectories $X^i_\EGO$ and $X^j_\EGO$ will be first embedded and encoded in parallel, without any data sharing in these stages.
The embedding network is a simple 2-layer MLP, and the subsequent encoding network is a 2-layer 2-head lightweight Transformer encoder, which only computes self-attention across each input trajectory sequence.
These light structures help describe what the neighbor-$j$'s observed trajectory roughly looks like from the ego-$i$'s perspective.
Unlike the heavier final predictor, these embedding and encoding networks could capture a \emph{quick snapshot} for the ego predictor to learn how ego $i$ would develop its interaction biases, rather than the over-smoothed or finely distributed trajectory representations to make the final prediction.
Denoting these embedding and encoding networks together as $E_\EGO$ (see its detailed network structures and settings in \TABLE{tab_symbols}) and feature dimension as $d$, we have features $\mathbf{f}^i_\EGO, \mathbf{f}^j_\EGO \in \mathbb{R}^{t_a \times d}$:
\begin{equation}
    \label{eq_ego_encode}
    \mathbf{f}^i_\EGO = E_\EGO \left(X^i_\EGO\right),\quad
    \mathbf{f}^j_\EGO = E_\EGO \left(X^j_\EGO\right).
\end{equation}

Wong \ETAL~\cite{wong2025reverberation} propose a trainable time-reindexing bilinear operator, the reverberation transform, to learn agents' behavioral latencies for handling different trajectory-changing events between two given temporal periods.
This transform employs a bilinear structure, with the two trainable kernels applied to the outer product similarity of the temporal features.
Inspired by its bilinear form, to gather information from both trajectories (features) $\mathbf{f}^i_\EGO$ and $\mathbf{f}^j_\EGO$, the ego predictor uses two kernels to explicitly formulate ego biases, including the reverberation kernel $\mathbf{R}^j_\EGO \in \mathbb{R}^{t_a \times t_b}$ and the insight kernel $\mathbf{I}^i_\EGO \in \mathbb{R}^{t_a \times K_I}$.
Here, similar to the original reverberation transform, kernel $\mathbf{R}^j_\EGO$ learns how this neighbor-$j$'s observed trajectory may naturally grow during the prediction period (the next $t_b$ steps).
Its main concern is to learn potential temporal latencies for the neighbor $j$ to handle its own interactions and to plan its future trajectories.
This process is considered to be objective, without any biases.
Given a 3-layer MLP, named $R_\EGO$ (see \TABLE{tab_symbols}), this reverberation kernel is learned by
\begin{equation}
    \mathbf{R}^j_\EGO = R_\EGO \left(\mathbf{f}^j_\EGO\right) \in \mathbb{R}^{t_a \times t_b}.
\end{equation}

In contrast, the proposed insight kernel $\mathbf{I}^i_\EGO \in \mathbb{R}^{t_a \times K_I}$ is the direct container to learn and anchor biased future assumptions produced by the ego $i$, therefore recombining and adjusting the objective temporal features $\mathbf{f}^j_\EGO$ according to the subjective $\mathbf{f}^i_\EGO$.
Specifically, this kernel is supposed to be shared for the same ego $i$ to consider all of its neighbors, even ego $i$ itself.
Given the number of insights $K_{I}$ and a similar 3-layer MLP $I_\EGO$ (\TABLE{tab_symbols}), we have
\begin{equation}
    \mathbf{I}^i_\EGO = I_\EGO \left(\mathbf{f}^i_\EGO\right) \in \mathbb{R}^{t_a \times K_I}.
\end{equation}

Then, similar to the original reverberation transform \cite{wong2025reverberation}, kernels $\mathbf{R}^j_\EGO$ and $\mathbf{I}^i_\EGO$ act on the similarity matrix of the neighbor representation $\mathbf{f}^j_\EGO$ via matrix multiplication to simulate this asymmetric rehearsal process.
Here, it should be noted that these asymmetric rehearsals appear explicitly as multimodal predictions of the ego predictor.
Although existing graph-based methods may have the ability to model asymmetric social interactions between agents, they mostly operate implicitly at the feature level for subsequent decoding networks, let alone addressing the asymmetric ego bias that grows beyond social interactions.
For the neighbor $j$, those $K_I$ ego biases captured within the $\mathbf{I}^i_\EGO$ are immediately applied to directing ego-$i$'s $K_I$ corresponding biased predictions.
For any $n$th ($n = 1, 2, ..., d$) feature dimension, the ego-$i$-biased neighbor-$j$'s trajectory representation $\overline{\mathbf{F}}^{j \leftarrow i}_\EGO \in \mathbb{R}^{K_I \times t_b \times d}$ is obtained by
\begin{equation}
    \label{eq_ego_bias_representation}
    \left(\overline{\mathbf{F}}^{j \leftarrow i}_\EGO\right)_{::n} = \left(\mathbf{I}^i_\EGO\right)^\top \left(\mathbf{F}^j_\EGO\right)_{::n} \mathbf{R}^j_\EGO,
\end{equation}
where
\begin{equation}
    \label{eq_ego_outer}
    \left(\mathbf{F}^j_\EGO\right)_{::n} = \left(\mathbf{f}^j_\EGO\right)_{:n} \left({\mathbf{f}^j_\EGO}\right)^\top_{:n} \in \mathbb{R}^{t_a \times t_a}.
\end{equation}
Here, $_{::n}$ and $_{:n}$ denote tensor slicing $_{:, :, n}$ and $_{:, n}$.

\NEWHALFLINE\subsubsection{Interpretations of Ego Biases}
\label{sec_method_interpretations}

Although \EQUA{eq_ego_bias_representation,eq_ego_outer} show similar forms to the reverberation transform \cite{wong2025reverberation}, this ego-biased feature-level modulation presents its own implications for learning and forecasting any ego-$i$'s biased interaction considerations.
The most significant difference is that both ego $i$ and neighbor $j$ are involved in contributing to forecasting neighbor-$j$'s future trajectories, while only the target agent $j$ (as well as its interaction context) has been directly considered in the original reverberation transform.
It should be noted that our referred ``involve'' indicates that a specific agent ($i$) has the ability to make direct interventions or modifications to the prediction-target's ($j$'s) final predictions.
Such subjective interventions or modifications are not equivalent to the passive social interactions encountered when forecasting.
Though interactions do have the ability to change prediction-target's trajectories, the participants as well as the final trajectory changes are mostly perceived and observed identically by all other neighbors, remaining unchanged regardless of variations in the ``observer''.

Differently, in \EQUA{eq_ego_bias_representation}, predictions for the same neighbor $j$ could be different depending on the ego $i$, or more specifically, depending on the insight kernel $\mathbf{I}^i_\EGO$.
In detail, representation $\overline{\mathbf{F}}^{j \leftarrow i}_\EGO$ is obtained after the left-multiplication of $i$'s insight kernel $\mathbf{I}^i_\EGO$ and the right-multiplication of $j$'s reverberation kernel $\mathbf{R}^j_\EGO$ on the symmetric sequential similarity $\mathbf{F}^j_\EGO \in \mathbb{R}^{t_a \times t_a \times d}$.
On each feature dimension, kernel $\mathbf{R}^j_\EGO$ recombines columns ($t_a$ observation steps) in the similarity matrix, learning the combination coefficients to represent latency dynamics of trajectory events during observation.
This product can be considered as ego-independent since only $j$'s observations have been considered during its computation.
After that, ego-$i$'s biases explicitly participate in the row transform as the insight kernel $\mathbf{I}^i_\EGO$, performing $K_I$ distinct combinations to alter these $t_b$ ego-independent feature products.
Therefore, unlike the reverberation transform, $\mathbf{I}^i_\EGO$ directly controls how the final representation would be shaped within a single multiplication, achieving the controllable multimodal biased ego prediction.

\NEWHALFLINE\subsubsection{Rehearsal Trajectories \& Ego Loss}

The ego predictor uses a single-layer no-activation MLP (denoted as $D_\EGO$, see \TABLE{tab_symbols}) to obtain its predicted ego-biased trajectories.
We use trajectory variable $\hat{Y}^{j \leftarrow i}_\EGO$ to denote this output.
Formally,
\begin{equation}
    \label{eq_ego_predictor_output}
    \hat{Y}^{j \leftarrow i}_\EGO = D_\EGO \left(\overline{\mathbf{F}}^{j \leftarrow i}_\EGO\right) \in \mathbb{R}^{K_I \times t_b \times 2}.
\end{equation}
Here, we have $\hat{Y}^{j \leftarrow i}_\EGO = \hatbf{Y}^{j \leftarrow i}_b$ when training from periods $\mathcal{T}_a$ to $\mathcal{T}_b$, and $\hat{Y}^{j \leftarrow i}_\EGO = \hatbf{Y}^{j \leftarrow i}_d$ (\IE, our defined \emph{short-term future rehearsals} or simply referred to as \emph{rehearsal trajectories}) when running the regular forecasting from periods $\mathcal{T}_c$ to $\mathcal{T}_d$.

For each ego-neighbor pair $(i, j)$, the ego predictor is trained using the supervision of the \textbf{Ego Loss} $\ell_\EGO^{j \leftarrow i}$.
It is the \emph{best-of-$K_I$} $\ell_2$ loss, computed only from the observation period $\mathcal{T}_{h} = \mathcal{T}_a \cup \mathcal{T}_b$, aiming at encouraging the multimodality of ego-$i$'s biases in considering neighbor-$j$'s trajectories.
Formally,
\begin{equation}
    \label{eq_ego_loss}
    \ell_\EGO^{j \leftarrow i} = \min_{k=1}^{K_I} \frac{1}{t_b} \sum_{t=1}^{t_b} \left\|
        \left(\hatbf{Y}^{j \leftarrow i}_b\right)_{k, t, :} - \left(\mathbf{X}^j_b\right)_{t, :}
    \right\|_2.
\end{equation}
Here, $_{k, t, :}$ and $_{t, :}$ denote the tensor slicing.

Note that the ego predictor is trained together with the final predictor, without further separation of trainable variables but sharing the same final loss function.
This selection is due to the consideration that each ego-$i$'s interaction and trajectory biases will be finalized in $i$'s own future trajectories rather than these short-term rehearsals of all its neighbors.
This joint training is also designed to maintain the consistency of these ego-biases captured in both the ego predictor and the final predictor, ensuring that the conditioning of final predictor can be correctly routed to different ego biases.
See \EQUA{eq_final_loss}.

\subsection{Bias Conditioning}
\label{sec_method_conditioning}

We now formulate how the final predictor is conditioned on the forecasted short-term future rehearsals $\hatbf{Y}^{j \leftarrow i}_d$ (for any ego-neighbor pair $(i, j)$) forecasted by the ego predictor.
As shown in \FIG{fig_method_horizons} (c), the ego predictor utilizes the last $t_a$ time steps ($\mathcal{T}_c$ in \EQUA{eq_ego_periods}) of the entire observation period $\mathcal{T}_h$ to predict trajectories for the first $t_b$ steps ($\mathcal{T}_d$) in the final prediction period $\mathcal{T}_f$.
Differently, the final predictor aims to predict trajectories over the entire prediction period $\mathcal{T}_f$ (with $t_f$ steps), using observations on the entire observation period $\mathcal{T}_h$ ($t_h$ steps).
In other words, these prediction periods satisfy
\begin{align}
    \mathcal{T}_c &\subset \mathcal{T}_h, \quad \max \mathcal{T}_c = \max \mathcal{T}_h, \\
    \mathcal{T}_d &\subset \mathcal{T}_f, \quad \min \mathcal{T}_d = \min \mathcal{T}_f.
\end{align}

In most trajectory prediction approaches, their predictors (our mentioned final predictor) typically use both ego-$i$'s observed trajectory $\mathbf{X}_h^i \in \mathbb{R}^{t_h \times 2}$ and all of its neighboring agents' trajectories $\{\mathbf{X}^j_h | j \in \mathcal{N}, j \neq i\}$ to infer potential social interactions and make final predictions for each ego $i$ (or for all the agents as egos simultaneously).
However, ego-$i$'s own unique biases for considering social interactions or making trajectory plans have not been explicitly considered and encoded when forecasting, manifesting as symmetric or exchangeable indices of egos' or neighbors' representations within these networks.
The ego predictor is then proposed to provide direct trajectory-level ego biases as the explicit condition for considering interactions and future trajectories from the ego-$i$'s point of view, simulating how each ego $i$ would direct all of its neighbors to play the \emph{dress rehearsal} of the upcoming interactions within its brain and finalize ego-$i$'s own future trajectories.

Before starting, we first construct full rehearsal trajectories for any ego-$i$'s neighbor $j$, directed by the ego $i$ itself.
For any $j \in \mathcal{N}$ and the $k$th $(k=1, ..., K_I)$ short-term rehearsal $\left(\hatbf{Y}^{j \leftarrow i}_d\right)_{k::}$, we denote its $k$th \emph{full rehearsal trajectory} as
\begin{equation}
    \label{eq_rehearsal_concat}
    \left(\hatbf{X}^{j \leftarrow i}_r\right)_{k::} = \left(\begin{matrix}
        \mathbf{X}^j_h \\
        \makebox[5em]{\dotfill} \\
        \left(\hatbf{Y}^{j \leftarrow i}_d\right)_{k::}
    \end{matrix}\right) \in \mathbb{R}^{(t_h + t_b) \times 2}.
\end{equation}
Here, $\hatbf{X}^{j \leftarrow i}_r \in \mathbb{R}^{K_I \times (t_h + t_b) \times 2}$.
Then, as illustrated in \FIG{fig_method_horizons} (d), when forecasting trajectories for ego $i$, the set of \emph{Biased Observations} $\mathcal{X}^i_\BIASED$ is used as the basic input to condition the final predictor, replacing the original $t_h$-step observations:
\begin{equation}
    \label{eq_biased_observations}
    \mathcal{X}^i_\BIASED = \left\{\hatbf{X}^{i \leftarrow i}_r\right\} \cup \left\{\hatbf{X}^{j \leftarrow i}_r \right|\left.\vphantom{\hatbf{X}^{j \leftarrow i}_r} j \in \mathcal{N}, j \neq i\right\}.
\end{equation}
This directly ensures that a set of unique trajectories $\mathcal{X}^i_\BIASED$ will be prepared when learning and forecasting each ego-$i$'s biased future trajectories within the final predictor, rather than forcing the network to capture those minor differences from the symmetric or isotropic features during the gradient descent.
Such additional $t_b$ rehearsal steps in the biased observations could also bring extra capabilities for learning the alignments or corrections from rehearsals to actual future trajectories, which further guides the training of the ego predictor.

Considering that Transformer \cite{vaswani2017attention} is still the main choice of backbone prediction models, these biased observations will be used to condition final predictions from the direct trajectory-level (as targets for evaluating cross attention) and the feature-level (as keys or queries) simultaneously.

\NEWHALFLINE\subsubsection{Feature-Level Conditioning}

Due to the use of the insight kernel $\mathbf{I}^i_\EGO$ with $K_I$ columns, each concatenated full rehearsal trajectory $\hatbf{X}^{j \leftarrow i}_r$ contains $K_I$ multimodal combinations of biased trajectories.
Not all of these rehearsals are supposed to represent the biased trajectory thoughts or interaction conditions.
Instead, we use a feature-wise max-pooling strategy to simultaneously take into account all \emph{Notable or Activated} feature units across all $K_I$ rehearsals.
Denoting the embedding and encoding networks in the final predictor as $E_\FINAL$ (with $d'$ output feature units, larger than the $d$ used in the ego predictor in \EQUA{eq_ego_encode}), the representation of all rehearsals $\mathbf{f}^{j \leftarrow i}_r$ and the final conditioning feature $\mathbf{f}^{j \leftarrow i}_\FINAL$ are computed as
\begin{align}
    \label{eq_method_maxpool}
    \mathbf{f}^{j \leftarrow i}_r &= E_\FINAL \left(\hatbf{X}^{j \leftarrow i}_r\right) \in \mathbb{R}^{K_I \times (t_h + t_b) \times d'}, \\
    \label{eq_method_condition}
    \mathbf{f}^{j \leftarrow i}_\FINAL &= \max_{k=1}^{K_I} \left(\mathbf{f}^{j \leftarrow i}_r\right)_{k::} \in \mathbb{R}^{(t_h + t_b) \times d'}.
\end{align}
When forecasting for the ego $i$, feature $\mathbf{f}^{j \leftarrow i}_\FINAL$ will be used to replace all these original non-biased ones in the final predictor to compute either further social interactions or cross-attentions in Transformer layers, providing unique asymmetric context for considering ego-$i$'s trajectory and interaction biases.

\NEWHALFLINE\subsubsection{Trajectory-Level Conditioning}

Unlike the above biased features $\mathbf{f}^{j \leftarrow i}_\FINAL$, the trajectory-level conditioning cares more about how all these ego rehearsals are overall distributed.
It provides the basis for evaluating all biased rehearsals of the same ego $i$, mostly playing the role of a reference or target trajectory (the \emph{value}) for computing cross attention.
Formally,
\begin{equation}
    \label{eq_method_mean}
    \mathbf{X}^{j \leftarrow i}_\FINAL = \mathbb{E}_{k} \left(\hatbf{X}^{j \leftarrow i}_r\right)_{k::} \in \mathbb{R}^{(t_h + t_b) \times 2}.
\end{equation}
Similar to the above feature-level conditioning, trajectory $\mathbf{X}^{j \leftarrow i}_\FINAL$ will be used to replace all other involved agents' observed trajectories when computing attention within Transformer layers, providing bases for learning ego biases.

Note that although the lengths of input trajectories have been modified to $t_h + t_b$ steps from the original $t_h$ observation steps, the prediction period of the final predictor is still $\mathcal{T}_f~(\| \mathcal{T}_f \| = t_f)$.
To suit these conditionings while maintaining clarity, we denote final predicted trajectories of the final predictor for the ego $i$ as $\hatbf{Y}^{i \leftarrow i}_\FINAL \in \mathbb{R}^{K \times t_f \times 2}$.
Here, $K$ represents the final number of generated multimodal trajectories.

\subsection{Training}

The proposed \MODEL~trajectory prediction model is constructed by integrating the ego predictor and its biased ego rehearsals into an existing trajectory prediction network (which serves as the final predictor).
In our experiments, we use the basic Transformer \cite{vaswani2017attention} and the \REVMODEL~\REVCITE~model as our final predictors.
Correspondingly, we have two \MODEL~variations, \MODEL-Rev and \MODEL-Trans\footnote{
    \MODEL-Rev will be treated as the main variation of \MODEL, while \MODEL-Trans only serves as a baisc ablation variation for analyzing the ego predictor.
    Unless otherwise specified, \MODEL~refers to the \MODEL-Rev by default. 
} (see settings in \SECTION{sec_exp_implementation}).
After applying both feature-level and trajectory-level conditioning, the following loss function will be used to train the ego predictor and the final predictor end-to-end together (for any ego $i$):
\begin{equation}
    \label{eq_final_loss}
    \ell^i = \ell^{i \leftarrow i}_\FINAL + \lambda \mathbb{E}_{j \in \mathcal{N}} \ell_\EGO^{j \leftarrow i},
\end{equation}
where $\lambda$ is a balancing parameter for the ego loss $\ell_\EGO^{j \leftarrow i}$ (see \EQUA{eq_ego_loss}), and the final loss $\ell^{i \leftarrow i}_\FINAL$ is defined as
\begin{equation}
    \ell^{i \leftarrow i}_\FINAL = 
        \min_{k=1}^{K} \frac{1}{t_f} \sum_{t=1}^{t_f} \left\|
        \left(\hatbf{Y}^{i \leftarrow i}_\FINAL\right)_{k, t, :} - \left(\mathbf{Y}^i_f\right)_{t, :}
    \right\|_2.
\end{equation}

\section{Experiments}

This section provides experiments and ablation studies to validate the proposed \MODEL~trajectory prediction model.
We will also provide further discussions to interpret how the ego predictor collaborates with the final predictor to achieve conditioned predictions using its forecasted ego-biased rehearsals.

\subsection{Experimental Settings}

\subsubsection{Datasets}
We use four public trajectory datasets to evaluate the proposed \MODEL~trajectory prediction model, including two pedestrian datasets (ETH-UCY and SDD), a vehicle dataset (nuScenes), and a game dataset (NBA):

(a) \textbf{ETH-UCY\cite{pellegrini2009youll,lerner2007crowds}} contains trajectories captured in pedestrian walking scenes.
It has 5 test scenes, corresponding to 5 training splits under the \emph{leave-one-out} \cite{alahi2016social,gupta2018social} strategy.
Following the common settings, we set $\{t_h, t_f, \Delta t\} = \{8, 12, 0.4\mbox{~(sec)}\}$.
Prediction metrics are reported in meters, under the \emph{best-of-20} stochastic evaluation \cite{gupta2018social}.

(b) \textbf{Stanford Drone Dataset (SDD) \cite{robicquet2016learning}} has 60 drone-captured trajectory scenes at Stanford.
It includes different types of agents, like pedestrians, bikers, and carts, posing more challenges for modeling interactions.
We randomly select 36 clips to train, 12 to validate, and the remaining 12 to test \SIMAUGCITE\GARDENCITE, using the common $\{t_h, t_f, \Delta t\} = \{8, 12, 0.4\}$ settings.
Metrics are reported in pixels, also under the \emph{best-of-20}.

(c) \textbf{nuScenes} \cite{caesar2019nuscenes} is a large-scale trajectory dataset, collected from various urban driving scenes.
It includes multiple kinds of finely labeled vehicles and pedestrians.
We use 550 scenes to train, 150 to validate, and 150 to test \cite{saadatnejad2022pedestrian}.
Following previous works \cite{yuan2021agentformer}, only vehicle trajectories are used, with the $\{t_h, t_f, \Delta t\} = \{4, 12, 0.5\}$ settings.
Metrics are reported in meters, under both \emph{best-of-5} and \emph{best-of-10} \cite{lee2022muse}.

(d) \textbf{NBA SportVU} (NBA)~\cite{linou2016nba} includes trajectories captured by SportVU tracking systems during NBA games, presenting intense interactions beyond collision avoidance.
Following previous works\cite{xu2022groupnet,xu2022remember}, we set $\{t_h, t_f, \Delta t\} = \{5, 10, 0.4\}$, and randomly select 50K samples, including 65\% for training and 25\%/10\% for test/validation.
Positions of players and basketballs are recorded in inches, while metrics are reported in meters, under the \emph{best-of-20} \cite{xu2022remember}.

\NEWHALFLINE
\subsubsection{Metrics}
We use the minimum Average/Final Displacement Error over $K$ trajectories (minADE$_K$/minFDE$_K$ or \emph{best-of-$K$} ADE/FDE) \cite{alahi2016social,pellegrini2009youll,gupta2018social} to evaluate prediction models.
For each agent $i$, given the stochastic prediction $\hatbf{Y}^i \in \mathbb{R}^{K \times t_f \times 2}$ and the groundtruth trajectory $\mathbf{Y}^i \in \mathbb{R}^{t_f \times 2}$, we have
\begin{align}
    \mbox{minADE}_{K}(i) &= \min_{k=1}^K \frac{1}{t_f} \sum_{t=1}^{t_f} \left\|
        \hatbf{Y}^i_{k,t,:} - \mathbf{Y}^i_{t,:}
    \right\|_2, \\
    \mbox{minFDE}_{K}(i) &= \min_{k=1}^K \left\|
        \hatbf{Y}^i_{k,t_f,:} - \mathbf{Y}^i_{t_f,:}
    \right\|_2.
\end{align}
The reported results are averaged for all test agents.
For clarity, we still refer to them as ``ADE/FDE'' for a fixed $K$.

\NEWHALFLINE
\subsubsection{Baselines}
The following recent state-of-the-art trajectory prediction approaches are selected as our baselines for comparisons:
\STAR\STARCITE~(2020),
\PECNET\PECNETCITE~(2020),
NMMP\cite{hu2020collaborative}~(2020),
\TRAJECTRONPP\TRAJECTRONPPCITE~(2020),
\YNET\YNETCITE~(2021),
\AGENTFORMER\AGENTFORMERCITE~(2021),
\SPECTGNN\SPECTGNNCITE~(2021),
MemoNet\cite{xu2022remember}~(2022),
GroupNet+NMMP\cite{xu2022groupnet}~(2022),
GroupNet+CVAE\cite{xu2022groupnet}~(2022),
\VMODEL\VCITE~(2022),
\MUSEVAE\MUSEVAECITE~(2022),
\MSRL\MSRLCITE~(2023),
\DICE\DICECITE~(2024),
\AFFLN\AFFLNCITE~(2024),
\SMEMO\SMEMOCITE~(2024),
\LGTRAJ\LGTRAJCITE~(2024),
\PPT\PPTCITE~(2024),
\UEN\UENCITE~(2024),
\MSTIP\MSTIPCITE~(2024),
\SCMODEL\SCCITE~(2024),
\SCPMODEL\SCPCITE~(2024),
\UPDD\UPDDCITE~(2024),
\EVMODEL\EVCITE~(2025),
\REMODEL\RECITE~(2025),
\REVMODEL\REVCITE~(2025),
and \SOPERMODEL\SOPERMODELCITE~(2025).

\NEWHALFLINE
\subsection{Implementation Details}
\label{sec_exp_implementation}
\MODEL~trajectory prediction model is trained end-to-end on one NVIDIA RTX 3090 GPU, using the Adam optimizer with the learning rate of 1e-4 and the batch size of 500 for up to 300 epochs.
For the ego predictor, we set $\{t_a, t_b\} = \{4, 4\}$ for the $t_h = 8$ datasets (ETH-UCY and SDD).
In contrast, we set $\{t_a, t_b\} = \{2, 2\}$ for the $t_h = 4$ nuScenes cases, while setting $\{t_a, t_b\} = \{3, 2\}$ for the $t_h = 5$ NBA.
For all these datasets, the number of insights ($K_I$ in the insight kernel) is set to 3, and the balancing parameter $\lambda$ for the ego loss is set to 0.6.
Also, following previous works \cite{wong2023another,wong2025reverberation}, the Haar transform is used on the input trajectories of the ego predictor to maximize its time-frequency joint representation capability.
The feature dimension of ego predictor layers (in \TABLE{tab_symbols}) is set to $d = 64$ ($d' = 128$ for the final predictor).
To speed up computation, the ego predictor is set to forecast the 5 closest neighbors for each ego agent, according to their Euclidean distances at the current observation step ($t = t_h$), while other neighbors will be forecasted by the linear-least-squares predictor that shares the same prediction settings.

For the final predictor, we select the \REVMODEL~\REVCITE~trajectory prediction model as the main variation of \MODEL~(\MODEL-Rev).
We keep the same model settings as reported in their original paper when training \MODEL, including feature dimensions, attention heads in Transformers, the number of Transformer layers, etc.
Also, this final predictor is responsible for generating $K$ stochastic predictions corresponding to the \emph{best-of-$K$} evaluation in each dataset.
All noise-sampling parts are the same as reported in their paper.
In addition, we have also constructed the vanilla Transformer~\cite{vaswani2017attention} variation, the \MODEL-Trans, to conduct further ablation studies and discussions.
Only 4 attention layers with 8 attention heads and the feature dimension $d'$ of 128 will be used to make final predictions (these settings are identical to Transformers used in the original \REVMODEL), without further consideration of social interactions and additional trajectory processes like trajectory transforms.
The code for all these variations, model weights or ablation weights, and detailed training settings are publicly available in our code repository.

\subsection{Comparisons to State-of-the-Arts}

\begin{table*}
    \caption{
        Comparisons with Recent State-of-the-art Methods on ETH-UCY (left) and SDD (right)
    }
    \label{tab_ethucysdd}
    \centering

    \begin{threeparttable}
        \begin{tabular}{@{} c @{\hspace{0.5em}} c @{}}
            \begin{tabular}{lcccccccccccccc}
                \toprule
                Method (ETH-UCY) & eth $\downarrow$ & hotel $\downarrow$ & univ $\downarrow$ & zara1 $\downarrow$ & zara2 $\downarrow$ & Average $\downarrow$ \\

                \midrule
                \MSTIP\MSTIPCITE~(2024)             & 0.39/0.57         & 0.13/0.22         & 0.24/\C{0.40}     & 0.20/0.34         & 0.17/0.29         & 0.22/0.36 \\
                \SMEMO\SMEMOCITE~(2024)             & 0.39/0.59         & 0.14/0.20         & \C{0.23/0.41}     & 0.19/0.32         & 0.15/0.26         & 0.22/0.35 \\
                \LGTRAJ\LGTRAJCITE~(2024)           & 0.38/0.56         & \C{0.11}/0.17     & \C{0.23}/0.42     & 0.18/0.33         & \C{0.14}/0.25     & 0.20/0.34 \\
                \MSRL\MSRLCITE~(2023)               & 0.28/0.47         & 0.14/0.22         & 0.24/0.43         & \C{0.17}/0.30     & \C{0.14/0.23}     & \C{0.19}/0.33 \\
                \PPT\PPTCITE~(2024)                 & 0.36/0.51         & \C{0.11/0.15}     & \C{0.22/0.40}     & \C{0.17}/0.30     & \C{0.12/0.21}     & 0.20/0.31 \\
                \EVMODEL\EVCITE~(2025)              & \C{0.25}/0.38     & \C{0.11/0.16}     & \C{0.23}/0.42     & 0.19/0.30         & \C{0.13}/0.24     & \C{0.18}/0.30 \\
                \AGENTFORMER\AGENTFORMERCITE~(2021) & 0.26/0.39         & \C{0.11/0.14}     & 0.26/0.46         & \C{0.15/0.23}     & \C{0.14}/0.23     & \C{0.18/0.29} \\
                \SCMODEL\SCCITE~(2024)              & \C{0.25}/0.38     & \C{0.12/0.14}     & \C{0.23}/0.42     & 0.18/0.29         & \C{0.13/0.22}     & \C{0.18/0.29} \\
                \SCPMODEL\SCPCITE~(2024)            & \C{0.25}/0.39     & \C{0.10/0.15}     & 0.24/0.42         & 0.18/\C{0.28}     & \C{0.13/0.22}     & \C{0.18/0.29} \\
                \YNET\YNETCITE~(2021)               & 0.28/\C{0.33}     & \C{0.10/0.14}     & 0.24/\C{0.41}     & \C{0.17/0.27}     & \C{0.13/0.22}     & \C{0.18/0.27} \\
                \UPDD\UPDDCITE~(2024)               & \C{0.22}/0.42     & 0.17/0.30         & \C{0.14/0.28}     & \C{0.16}/0.30     & \C{0.14}/0.31     & \C{0.17}/0.32 \\
                \REMODEL\RECITE~(2025)              & \C{0.23/0.35}     & \C{0.10/0.15}     & 0.24/\C{0.41}     & \C{0.17}/0.29     & \C{0.13/0.22}     & \C{0.17/0.28} \\ 
                \REVMODEL\REVCITE~(2025)            & \C{0.23/0.34}     & \C{0.10/0.15}     & \C{0.23/0.40}     & \C{0.17/0.28}     & \C{0.13/0.22}     & \C{0.17/0.28} \\ 

                \midrule
                \MODEL~(Ours)                       & \C{0.23/0.35}     & \C{0.11/0.15}     & \C{0.23/0.40}     & \C{0.17/0.28}     & \C{0.12/0.22}     & \C{0.17/0.28} \\

                \bottomrule
            \end{tabular}

            &

            \begin{tabular}{lc}
                \toprule
                Method (SDD) & ADE/FDE $\downarrow$ \\

                \midrule
                \SPECTGNN\SPECTGNNCITE~(2021)   & 8.21/12.41 \\
                \SMEMO\SMEMOCITE~(2024)         & 8.11/13.06 \\
                \LGTRAJ\LGTRAJCITE~(2024)       & 7.80/12.79 \\
                \YNET\YNETCITE~(2021)           & 7.85/11.85 \\
                \UEN\UENCITE~(2024)             & 7.30/10.40 \\
                \PPT\PPTCITE~(2024)             & 7.03/10.65 \\
                \UPDD\UPDDCITE~(2024)           & 6.59/13.90 \\
                \EVMODEL\EVCITE~(2025)          & 6.57/10.49 \\
                \SCMODEL\SCCITE~(2024)          & 6.54/10.36 \\
                \SCPMODEL\SCPCITE~(2024)        & 6.44/10.22 \\
                \MUSEVAE\MUSEVAECITE~(2022)     & 6.36/11.10 \\
                \REMODEL\RECITE~(2025)          & \C{6.27/10.02} \\
                \REVMODEL\REVCITE~(2025)        & \C{6.14/9.87} \\
                
                \midrule
                \MODEL~(Ours)                   & \C{6.11/9.84} \\

                \bottomrule
            \end{tabular}
        \end{tabular}

        \begin{tablenotes}[flushleft]
            \footnotesize
            \item[]
                Metrics are reported as ``ADE/FDE'' (\emph{best-of-20}).
                Lower metrics indicate better prediction performance.
                \C{Blue values} denote the best 3 results on each set.
        \end{tablenotes}
    \end{threeparttable}
\end{table*}

\begin{table}[tbp]
    \caption{
        Comparisons with Recent State-of-the-arts on nuScenes
    }
    \label{tab_nuscenes}
    \centering
    \begin{threeparttable}
        \begin{tabular}{lcc}
            \toprule
            Method (nuScenes) & \emph{best-of-5} $\downarrow$ & \emph{best-of-10} $\downarrow$ \\

            \midrule
            \TRAJECTRONPP\TRAJECTRONPPCITE~(2020)   & 3.14/7.45 & 2.46/5.65 \\
            \YNET\YNETCITE~(2020)                   & 2.46/5.15 & 1.88/3.47 \\
            \SOPERMODEL\SOPERMODELCITE~(2025)       & 2.21/4.58 & 1.79/3.40 \\
            \AGENTFORMER\AGENTFORMERCITE~(2021)     & 1.86/3.89 & 1.45/2.86 \\
            \DICE\DICECITE~(2024)                   & 1.76/3.70 & 1.44/2.67 \\
            \AFFLN\AFFLNCITE~(2024)                 & 1.83/3.78 & 1.32/2.73 \\
            \EVMODEL\EVCITE~(2025)                  & 1.46/3.18 & 1.15/2.37 \\
            \SCMODEL\SCCITE~(2024)                  & 1.44/3.10 & 1.13/2.30 \\
            \MUSEVAE\MUSEVAECITE~(2022)             & 1.38/2.90 & 1.09/2.10 \\
            \REMODEL\RECITE~(2025)                  & \C{1.26/2.72} & \C{1.03/2.12} \\
            \REVMODEL\REVCITE~(2025)                & \C{1.25/2.73} & \C{1.00/2.03} \\

            \midrule
            \MODEL~(Ours) & \C{1.24/2.71} & \C{0.99/2.02} \\

            \bottomrule
        \end{tabular}

        \begin{tablenotes}[flushleft]
            \footnotesize
            \item[]
                Metrics (``ADE/FDE'') are reported under \emph{best-of-5} and \emph{best-of-10}.
                \C{Blue values} denote the best 3 results under each setting.
        \end{tablenotes}
    \end{threeparttable}
\end{table}

\begin{table}[tbp]
    \caption{
        Comparisons with Recent State-of-the-arts on NBA
    }
    \label{tab_nba}
    \centering
    \begin{threeparttable}
        \begin{tabular}{lcc}
            \toprule
            Method (NBA) & 2.0s $\downarrow$ & 4.0s $\downarrow$ \\

            \midrule
            \STAR\STARCITE~(2020) & 0.77/1.28 & 1.26/2.04 \\
            \PECNET\PECNETCITE~(2020) & 0.96/1.69 & 1.83/3.41 \\
            NMMP\cite{hu2020collaborative}~(2020) & 0.70/1.11 & 1.33/2.05 \\
            MemoNet\cite{xu2022remember}~(2022) & 0.71/1.14 & 1.25/1.47 \\
            GroupNet+NMMP\cite{xu2022groupnet}~(2022) & 0.69/1.08 & 1.25/1.80 \\
            GroupNet+CVAE\cite{xu2022groupnet}~(2022) & \C{0.62}/0.95 & \C{1.13}/1.69 \\
            \VMODEL\VCITE~(2022) & 0.69/0.96 & 1.28/1.68 \\
            \EVMODEL\EVCITE~(2025) & 0.68/0.93 & 1.26/1.64 \\
            \SCMODEL\SCCITE~(2024) & 0.67/0.90 & 1.18/1.46 \\
            \SCPMODEL\SCPCITE~(2024) & 0.65/\C{0.86} & 1.14/\C{1.37} \\
            \REMODEL\RECITE~(2025) & \C{0.60/0.78} & \C{1.12/1.38} \\
            
            \midrule
            \MODEL~(Ours) & \C{0.61/0.76} & \C{1.11/1.34} \\

            \bottomrule
        \end{tabular}

        \begin{tablenotes}[flushleft]
            \footnotesize
            \item[]
                Metrics (``ADE/FDE'') are reported under \emph{best-of-20} for two different prediction horizons (2.0s ($t_f = 5$) and 4.0s ($t_f = 10$)).
                \C{Blue values} denote the best 3 results under each setting.
        \end{tablenotes}
    \end{threeparttable}
\end{table}

We first compare \MODEL~with other recent state-of-the-art trajectory prediction methods to validate the overall quantitative performance.
Metrics are reported in \TABLE{tab_ethucysdd,tab_nuscenes,tab_nba}.
It can be seen that the proposed \MODEL~model (indicate the \MODEL-Rev by default) obtains comparable quantitative performance across multiple datasets, not only limited to pedestrian scenes (ETH-UCY, SDD) but traffic scenes (nuScenes) and the more challenging NBA scene.
For example, in \TABLE{tab_ethucysdd}, \MODEL~outperforms current state-of-the-art \REMODEL~and \REVMODEL~by up to 2.6\% ADE and 1.8\% FDE, respectively.
It also presents greater prediction performance on the more challenging nuScenes (\TABLE{tab_nuscenes}) and NBA (\TABLE{tab_nba}) datasets, including the 3.9\%/4.7\% better ADE/FDE on nuScenes and the 2.9\% better FDE on NBA (4.0s), compared to the strong \REMODEL~model.
Notably, although these gains are not dramatic, they are still consistent across datasets and more visible in longer-horizon (NBA results in \TABLE{tab_nba}) or larger-K settings (\emph{best-of-10} nuScenes results in \TABLE{tab_nuscenes}).
Given the limitations and current bottleneck states of these datasets, particularly the annotation noise in certain datasets (such as the eth and hotel in ETH-UCY), such consistent performance improvements actually demonstrate the capability of the proposed \MODEL~in handling different trajectory scenarios.
We have provided detailed ablation comparisons (with greater numerical precision) in \SECTION{sec_ablation} to further discuss each component in the proposed \MODEL~model.

\subsection{Ablation Studies}
\label{sec_ablation}

\begin{table*}
    \caption{
        Ablation Results of \MODEL-Trans (a1 to a5) and \MODEL-Rev (a6 to a13) on ETH-UCY and SDD
    }
    \label{tab_ablation_ethucy}
    \centering

    \begin{threeparttable}
        \begin{tabular}{lcccccccccccccc}
            \toprule
            \multirow{2}{*}[-0.7ex]{ID} &
            \multicolumn{4}{c}{Ego Predictor} &
            \multirow{2}{*}[-0.7ex]{Final Predictor} &
            \multirow{2}{*}[-0.7ex]{eth} &
            \multirow{2}{*}[-0.7ex]{hotel} &
            \multirow{2}{*}[-0.7ex]{univ} &
            \multirow{2}{*}[-0.7ex]{zara1} &
            \multirow{2}{*}[-0.7ex]{zara2} &
            \multirow{2}{*}[-0.7ex]{SDD} \\
            \cmidrule(lr){2-5} & Mod. & $K_I$ & $t_a$ & $t_b$\\

            \midrule
            a1 & None & -- & -- & -- & Transformer                  & 0.832/1.662           & 0.257/\C{0.441}       & 0.772/1.396           & 0.483/0.972           & 0.386/0.740           & 17.445/33.368 \\
            a2 & Linear & -- & 4 & 4 & Transformer                  & \C{0.605}/\C{1.234}   & \C{0.246}/0.447       & 0.620/1.313           & \C{0.401}/0.881       & 0.312/\C{0.682}       & 14.383/29.866 \\
            a3 & $j \leftarrow j$ & 3 & 4 & 4 & Transformer         & 0.615/\CC{1.232}      & 0.258/0.502           & 0.616/\C{1.304}       & 0.403/\C{0.878}       & \C{0.311}/0.691       & 14.345/\C{29.817} \\
            a4 & $j \leftarrow i$ & 1 & 4 & 4 & Transformer         & 0.633/1.257           & 0.249/0.445           & \C{0.614}/1.315       & 0.404/0.892           & 0.312/0.697           & \C{14.338}/29.970 \\
            a5 & $j \leftarrow i$ & 3 & 4 & 4 & Transformer         & \CC{0.593}/1.279      & \CC{0.243}/\CC{0.436} & \CC{0.607}/\CC{1.295} & \CC{0.400}/\CC{0.876} & \CC{0.309}/\CC{0.677} & \CC{14.327}/\CC{29.785} \\

            \midrule
            a6      & None              & -- & -- & -- & \REVMODEL  & \CC{0.229}/\CC{0.347} & \CC{0.105}/\C{0.154}  & 0.237/\C{0.409}       & 0.172/\C{0.283}       & 0.131/0.220           & \C{6.144}/\C{9.874} \\
            a7      & Linear            & -- & 4  & 4  & \REVMODEL  & 0.237/0.359           & 0.116/0.174           & 0.242/0.423           & 0.173/0.297           & 0.130/0.223           & 6.204/10.133 \\
            a8      & $j \leftarrow i$  & 1  & 4  & 4  & \REVMODEL  & 0.238/0.365           & \C{0.112}/0.165       & 0.237/0.410           & 0.171/0.294           & 0.127/0.218           & 6.225/10.056 \\
            a9      & $j \leftarrow i$  & 5  & 4  & 4  & \REVMODEL  & 0.236/0.367           & 0.113/0.168           & 0.237/0.414           & 0.172/0.298           & 0.130/0.224           & 6.238/10.078 \\
            a10     & $j \leftarrow i$  & 10 & 4  & 4  & \REVMODEL  & 0.238/0.367           & 0.118/0.177           & 0.240/0.415           & 0.173/0.298           & 0.129/0.222           & 6.245/10.102 \\
            a11     & $j \leftarrow i$  & 3  & 6  & 2  & \REVMODEL  & 0.234/0.360           & 0.114/0.164           & 0.236/0.412           & 0.173/0.296           & 0.129/0.224           & 6.189/10.052 \\
            a12     & $j \leftarrow i$  & 3  & 2  & 6  & \REVMODEL  & 0.273/0.416           & 0.124/0.170           & \CC{0.229}/0.415      & \CC{0.166}/0.291      & \CC{0.124}/\CC{0.217} & 6.217/10.177 \\
            a13     & $j \leftarrow j$  & 3  & 4  & 4  & \REVMODEL  & 0.240/0.372           & 0.115/0.172           & 0.241/0.418           & 0.171/0.297           & 0.130/0.226           & 6.234/10.136 \\
            full    & $j \leftarrow i$  & 3  & 4  & 4  & \REVMODEL  & \C{0.232}/\C{0.357}   & 0.114/\CC{0.149}      & \C{0.230}/\CC{0.401}  & \C{0.169}/\CC{0.285}  & \C{0.125}/\CC{0.217}  & \CC{6.114}/\CC{9.840} \\

            \bottomrule
        \end{tabular}

        \begin{tablenotes}[flushleft]
            \footnotesize
            \item[]
                Metrics (ADE/FDE) are reported under \emph{best-of-20}.
                ``Mod.'' denotes in which way the ego predictor forecasts trajectories, where ``None'' indicates the ego predictor is disabled, and ``Linear'' indicates the linear-least-squares ego predictor. $j \leftarrow j$ denotes no ego-bias is considered, while $j \leftarrow i$ otherwise.
                Top-2 results in each dataset are highlighted in \C{blue}, and the best results are \CC{underlined}.
        \end{tablenotes}
    \end{threeparttable}

\end{table*}

We have conducted ablation experiments on both ETH-UCY and SDD to further validate how each component works in the proposed \MODEL.
These ablation results are reported in \TABLE{tab_ablation_ethucy} (our reported results are averaged over 5 test runs).

\NEWHALFLINE\subsubsection{Minimal Verifications}

We first discuss minimal ablation variations (a1 to a5 in \TABLE{tab_ablation_ethucy}) to directly verify how the ego predictor, as well as its forecasted short-term future rehearsal trajectories, collaborates with the final predictor.
To minimize all other effects brought by the final predictor, these variations only use a simple Transformer backbone as the final predictor, without the further modeling of social interactions and stochastic capabilities.
This means that the ego predictor only provides rehearsals for considering ego-agents' own future trajectories, rather than biased social interactions.

We see that the prediction performance of the vanilla Transformer (variation a1) has been greatly improved by collaborating with the ego predictor, even for the simple non-biased linear ego predictor (variation a2).
It shows that a2 outperforms a1 by a notable 27.2\% and 25.8\% in ADE and FDE on eth, and also about 19.7\% ADE on univ.
This linear ego predictor only provides a non-biased short-term linear prediction for each agent, thereby extending the original input trajectories (observations) of the final predictor.
Under this setting, the final predictor is forced to learn the spatial-temporal residuals between these extended short-term linear rehearsals and their complete future trajectories, rather than inferring trajectories only from the observation period $\mathcal{T}_h$.
Such improvements verify that the concatenation in \EQUA{eq_rehearsal_concat} could bring considerable performance gains, even when using the simple Transformer as the final predictor.
In addition, by further introducing the biased ego predictor and the number of insights $K_I$, minimal variations \{a3, a4, a5\} also show considerable performance improvements over the base variation a1, indicating the direct effectiveness of the ego predictor even for a relatively simple final predictor (see analyses $K_I$ below).

\NEWHALFLINE\subsubsection{Verifications of Ego Biases}

In addition to the above-mentioned spatial-temporal input extension, the core concern of the proposed ego predictor is to provide biased ego rehearsals for the subsequent final predictor.
We have conducted two sets of variations, including minimal variations \{a2, a3, a5\} and \MODEL-Rev variations \{a7, a13, full\}, to verify how the ego biases affect or contribute to the final predictor.

The ``$j \leftarrow j$'' variations (a3 and a13) replace the interaction bias term, \IE, all $\hatbf{X}_r^{j \leftarrow i}$ in \EQUA{eq_biased_observations}, with the non-biased ones $\hatbf{X}_r^{j \leftarrow j}$.
It means that the learning of ego biases has been structurally disabled in these non-biased variations (a3 and a13), while other network components remain identical to the full variations (a5 and full).
\TABLE{tab_ablation_ethucy} shows that the absence of such ego biases leads to direct performance drops.
For example, variation a3 falls behind a5 by about 6.2\%/15.1\% in ADE/FDE on hotel.
Notably, on the other datasets, the performance of a3 is comparable to or even slightly better than that of a5 (3.7\% better FDE on eth while almost the same ADE/FDE on other sets).
This is more likely due to the Transformer used not containing any interaction-modeling component.
For these minimal variations, ego biases actually only act on the ego loss during the training of the ego predictor itself, and those biased rehearsals do not directly contribute to the subsequent final predictor (since the interaction bias $j \leftarrow i$ term in \EQUA{eq_biased_observations} has been short-circuited).
Nevertheless, some datasets still present an over 15\% performance gain, which verifies the significant role biases have played during training.

Unlike these minimal variations, the $j \leftarrow j$ variation a13 consistently underperforms the biased $j \leftarrow i$ full model, including a notable 15.4\% worse FDE on hotel, 4.8\%/4.2\% worse ADE/FDE on univ, 4.0\%/4.1\% on zara2, etc.
These differences effectively show how the interaction bias term contributes to the whole \MODEL~in conditioning predictions, simply by introducing this asymmetric, anisotropic term in \EQUA{eq_biased_observations}.
It is worth noting that both biased and non-biased ego predictor variations \{a13, full\} far outperform the linear ego predictor variation a7, which demonstrates that the quality of short-term rehearsals themselves may also influence the final predictor, regardless of whether they are biased or not.
See our discussions about how the short-term rehearsal trajectories reflect these ego biases in \SECTION{sec_discussions_biases}.

\NEWHALFLINE\subsubsection{The Number of Insights $K_I$}
\label{sec_ablation_K}

$K_I$ controls the number of output channels in \EQUA{eq_ego_bias_representation}, \IE, the number of biases for the ego agent to \emph{re-evaluate} future trajectories or interactions with different neighbors.
Comparing minimal variations \{a4, a5\} or \MODEL-Rev variations \{a8, full, a9, a10\} in \TABLE{tab_ablation_ethucy}, it can be seen that neither too few nor too many $K_I$ leads to better quantitative performance.
Although the $K_I = 1$ minimal variation a4 and the $K_I = 3$ a5 present similar FDEs ($\pm$2.5\%) on ETH-UCY and SDD, a4 still underperforms a5 by a notable 6.7\% in ADE on eth.
Similar results also exist for variations \{a8, full, a9, a10\}, where the $K_I = 10$ variation a10 falls behind the $K_I = 3$ full variation by about 4.3\% in ADE and 3.5\% in FDE on univ, as well as 2.7\% FDE on SDD.
Also, the $K_I = 1$ a8 has a 10.7\% worse FDE on hotel compared to the full variation, even though a8 has a slightly better ADE (0.002 meters) than it.

Here, we can infer from these comparisons that the number of insights directly controls how the ego predictor collaborates with the final predictor.
When $K_I$ is too small, the stochastic capability of the ego predictor might be limited, making it difficult to distinguish different ego-agents' ego biases and to predict stable or distinguishable short-term rehearsal trajectories.
Further, since rehearsals are used as part of the input trajectories for the final predictor (\EQUA{eq_rehearsal_concat}), this could further make the whole network unable to fit these trajectories, even in the relatively simple hotel scene.
On the contrary, a larger $K_I$ may assign more gradient shares (\EQUA{eq_final_loss}) to the ego predictor, shrinking the further interaction-modeling capabilities of the final predictor, making it perform worse in dense scenes like univ.
We have provided further discussions to verify this thought.
See \SECTION{sec_discussion_maxpool}.

\NEWHALFLINE\subsubsection{Prediction Horizon $\{t_a, t_b\}$}

As elaborated in \SECTION{sec_method_horizon}, the ego predictor is trained on the prediction period $\mathcal{T}_h$ to simulate how humans perceive their interaction context while avoiding the data leakage from the prediction period $\mathcal{T}_f$.
This means that observation and prediction periods of the ego predictor are both limited to separate all $t_h$ steps in $\mathcal{T}_h$.
We conduct variations \{a11, full, a12\} in \TABLE{tab_ablation_ethucy} to validate how three different $t_a/t_b$ separations affect the overall prediction performance, under the $t_h = 8$ common setting.
It can be seen that the shorter prediction period variation a11 ($\{t_a, t_b\} = \{6, 2\}$) performs worse than the full model across all these datasets, like 10.1\% worse FDE on hotel and 2.6\% worse ADE on univ.
Differently, we notice an interesting phenomenon that the $\{t_a, t_b\} = \{2, 6\}$ longer prediction variation a12 achieves even slightly better ADE compared to the full \MODEL-Rev model, including 0.001 meters of ADE on univ, 0.003m on zara1, and 0.001m on zara2.
However, as a trade-off, a12 underperforms the full model for about 3.5\% FDE on univ and 2.1\% FDE on zara1.

We can infer that a longer prediction horizon ($t_b$) could provide the final predictor with more prior information, thereby leading to some improvements in the average prediction performance.
Notably, for the ego predictor, a longer prediction horizon leads to a shorter observation period, which means that the amount of information contained in the observation and the prediction period is not equivalent, and thus cannot effectively constrain the degrees of freedom in the forecasted rehearsal trajectories.
This finally results in an increased prediction error for the ego predictor, especially near the end of the prediction horizon, which consequently leads to the significant FDE drops (a12).
On the contrary, a shorter prediction horizon could not bring temporally rich ego biases to the final predictor.
In these cases, the ego predictor may forecast almost the same predictions for each agent, regardless of our concerned ego biases.
Such predictions could even confuse the whole network via the mixed timeline in \EQUA{eq_rehearsal_concat}, making it perform worse than the base variation a6 in \TABLE{tab_ablation_ethucy}.
Therefore, the $\{t_a, t_b\} = \{4, 4\}$ full model captures the equilibrium for balancing both overall and final prediction performance.

\subsection{Efficiency Analyses}

\begin{table}[tbp]
    \caption{
        Comparisons of Parameters and Time Costs
    }
    \label{tab_efficiency}
    \centering
    \begin{threeparttable}
        \setlength{\tabcolsep}{4pt}
        \begin{tabular}{ccccccccccccccc}
            \toprule
            \multirow{2}{*}[-0.7ex]{ID} & 
            \multirow{2}{*}[-0.7ex]{\makecell{Ego\\Predictor}} &
            \multirow{2}{*}[-0.7ex]{\makecell{Final\\Predictor}} &
            \multirow{2}{*}[-0.7ex]{Parameters} & 
            \multicolumn{4}{c}{Time Costs (ms)} \\
            \cmidrule(lr){5-7} &&&& 1 & 100 & 500 \\

            \midrule
            e1 & None   & Transformer & 1,921,166 (base) & 9 & 11 & 16 \\
            e2 & Linear & Transformer & 1,921,166 (+0.0\%) & 11 & 15 & 24 \\
            e3 & Full   & Transformer & 2,113,271 (+10.0\%) & 26 & 33 & 58 \\

            \midrule
            e4 & None & \REVMODEL & 3,008,636 (base) & 31 & 63 & 178 \\
            e5 & Linear & \REVMODEL & 3,008,636 (+0.0\%) & 44 & 88 & 251 \\
            e6 & Full & \REVMODEL & 3,200,741 (+6.4\%) & 78 & 134 & 290 \\
                
            \bottomrule
        \end{tabular}

        \begin{tablenotes}[flushleft]
            \footnotesize
            \item[]
                Results are obtained on one Mac mini M1 (8GB, 2020).
                Numbers below ``time costs'' indicate different batch sizes used.
                For \REVMODEL-based variations, 20 trajectories will be generated within one model inference.
        \end{tablenotes}
    \end{threeparttable}
\end{table}

The \MODEL~requires additional forecasting of rehearsal trajectories beyond the full-functional final predictor.
We now discuss whether such ego predictor affects its prediction efficiency, taking both variations \MODEL-Trans and \MODEL-Rev as examples.
All computations are conducted on one Apple M1 Mac mini (2020, 8GB Memory) for a fair comparison.
As reported in \TABLE{tab_efficiency}, \MODEL~still keeps considerable model efficiency on the resource-constrained computation platform.
Compared to the base models (variations e1 and e4), the ego predictor only requires less than 10.0\% extra parameters in the full \MODEL~models e3 and e6.
It is worth noting that the introduction of the proposed ego predictor causes the final predictor's prediction period to increase to $\mathcal{T}_h \cup \mathcal{T}_d$.
For the ETH-UCY test scene we selected, the final predictor must extract features from $\|\mathcal{T}_h\| + \|\mathcal{T}_d\| = 12$ trajectory frames for prediction, rather than the original $t_h = 8$.
Therefore, even the linear variations (e2 and e5) could lead to additional computations.
Nevertheless, when generating 20 predictions per agent, a single inference of the full \MODEL-Rev (e6) still takes only 78 ms with a batch size of 1.
Also, the number of trajectories generated can even reach approximately $20 \times 500 \times 1/0.290 \approx 34,000$ per second when the batch size is set to 500, particularly considering that this result is achieved on a unified-memory platform without a discrete GPU.

\section{Discussions}

This section discusses how components collaborate in the proposed \MODEL~model to achieve the goal of ego-bias-conditioned trajectory prediction beyond the only quantitative prediction performance.
We first visualize trajectories forecasted by the entire \MODEL~model and the ego predictor, exploring whether and how our focused ego biases exist and are formulated.
Then, by introducing a simple counterfactual validation and the quantitative metric \emph{Activation Rate}, we further try to provide interpretations on how ego biases finally have the ability to condition \MODEL~predictions.

\subsection{Visualized Model Predictions}

\begin{figure}[tbp]
    \centering
    \includegraphics[width=1.0\linewidth]{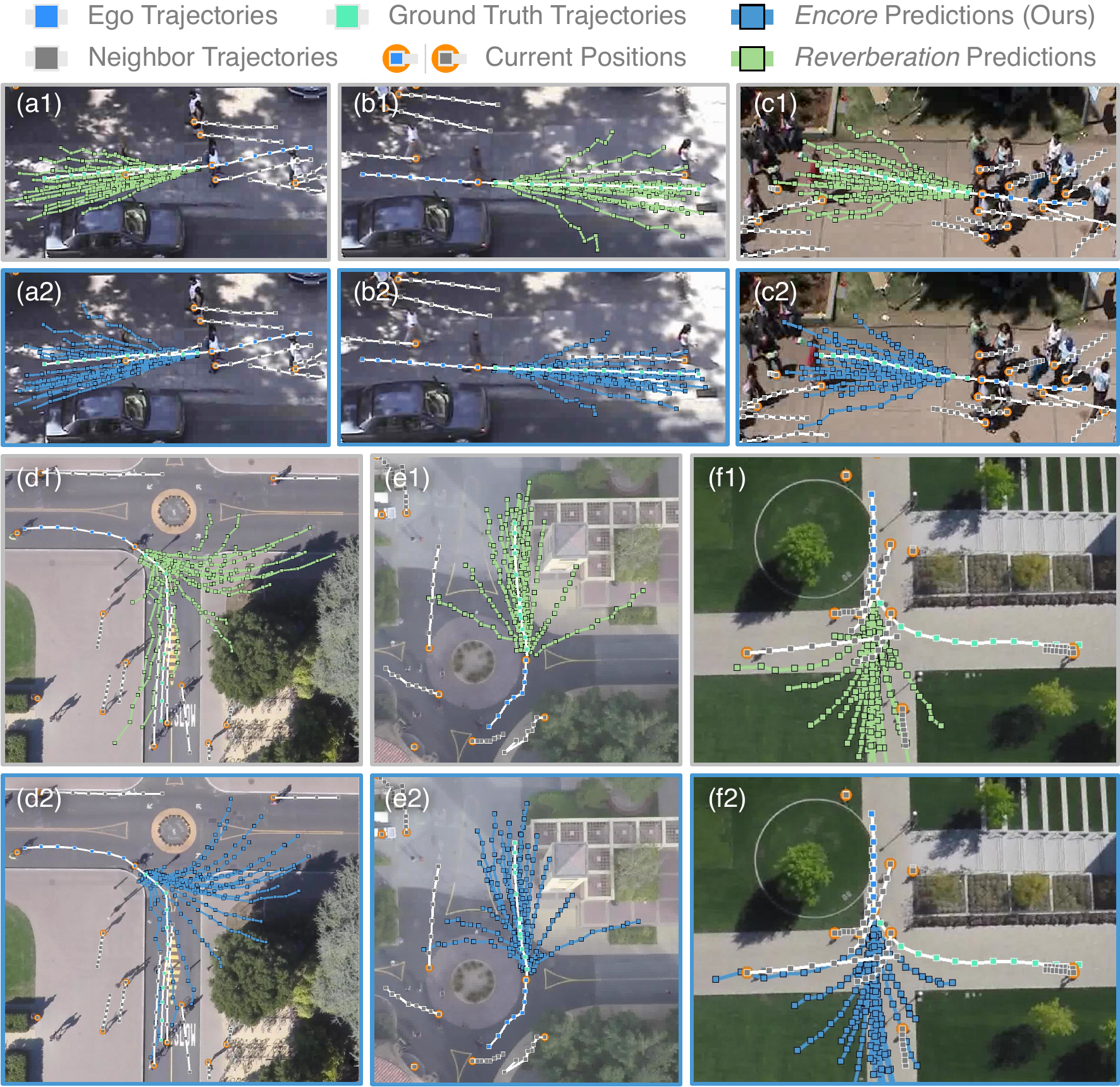}
    \caption{
        Visualized \MODEL~predictions and comparisons to the baseline \REVMODEL~\REVCITE~model's predictions in several example scenes.
    }
    \label{fig_exp_vis}
\end{figure}

\FIG{fig_exp_vis} visualizes \MODEL~predictions in several scenes, taking the \REVMODEL~\REVCITE~as the baseline.
In \FIG{fig_exp_vis} (a1), (b1), (c1), trajectories forecasted by the baseline model present \emph{homogenized} properties and multimodality patterns, even though these egos have notable trajectory differences and distinct interactive contexts.
Here, we observe that distinct prediction ``styles'' can be easily found in the \MODEL~predictions, although two models share similar backbone prediction networks (except for the bias-modeling and bias-conditioning components).

\NEWHALFLINE\subsubsection{Ego Biases in Predictions}

For example, some predictions in \FIG{fig_exp_vis} (a2) exhibit distinct trends of high-speed linear motion, maintaining the ego's overall movement direction throughout the observation period.
Other predictions primarily offer alternative interaction options, resulting in curved trajectories with more varied speeds and spatial positions.
Differently, such predictions can be rarely seen in other prediction cases. Instead, predictions are generated to better suit either current interactive contexts or the own states of ego agents, \IE, our focused ego biases, rather than sharing almost all trajectory patterns as those forecasted \FIG{fig_exp_vis} in (a1), (b1), and (c1).
Also, we notice that \MODEL~could provide greater trajectories for those cases with more nonlinear motions, as visualized in \FIG{fig_exp_vis} (d2) and (e2).
Although the scene collision maps or segmentation maps are not included as an input to the network, \MODEL~is still able to take into account both ego agents' and their neighbors' states simultaneously and finally make predictions with clearer multimodal characteristics to bypass roundabouts, rather than the plainly divergent trajectories of the original \REVMODEL~in \FIG{fig_exp_vis} (d1) and (d2).
Such predictions could verify the overall effectiveness of the proposed \MODEL~model qualitatively.

\NEWHALFLINE\subsubsection{Failure Cases}

Notably, as illustrated in \FIG{fig_exp_vis} (f2), \MODEL~still has failure predictions (where the forecasted trajectories completely fail to cover the groundtruth trajectory).
In this relatively complex intersection scene, the ego agent first yields to agents approaching from other directions, and then accelerates to turn left according to the groundtruth trajectory.
Given the interactive context, predictions of \MODEL~suggest that this ego is more likely to turn right or just go straight, rather than suddenly turning left in the near future.
This means that \MODEL~may still struggle to handle such unexpected interactive events (especially since the observation of this ego exhibits a clear linear motion trend).
Nevertheless, compared to the predictions in \FIG{fig_exp_vis} (f1), \MODEL~could still provide one predicted trajectory for considering a slower left-turning trend among all 20 predictions, demonstrating its potential to overcome such kinds of failures.

\subsection{Discussions of Rehearsals}
\label{sec_discussions_biases}

\begin{figure*}[tbp]
    \centering
    \includegraphics[width=1.0\linewidth]{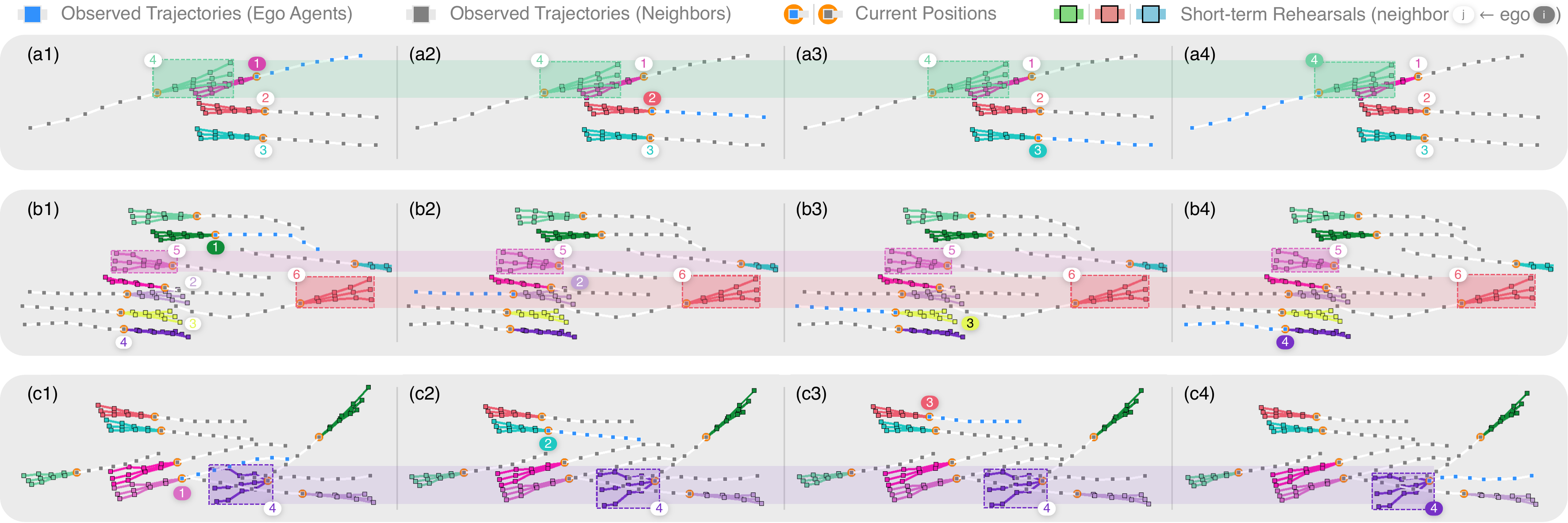}
    \caption{
        Comparisons of rehearsal trajectories $\hatbf{Y}_d^{j \leftarrow i} \in \mathbb{R}^{3 \times 4 \times 2}$ ($K_I = 3$) forecasted by the ego predictor in several ETH-UCY and SDD scenes.
        Predictions for different agents are distinguished by colors.
        The $i$th agent ($i = 1, 2, 3, 4$) will be considered as the ego agent in subfigures \{(a$i$), (b$i$), (c$i$)\} correspondingly.
        Predictions to be analyzed are highlighted in dashed boxes, covered with horizontal colorbars for reference (based on predictions in cases (a1), (b1), and (c1)).
    }
    \label{fig_exp_rehearsals}
\end{figure*}

The prediction pipeline of the \MODEL~model follows a \emph{rehearsal to encore} strategy.
During this process, the core consideration of its ego predictor is to learn to predict asymmetric, ego-biased short-term rehearsal trajectories (``\emph{rehearsals}'' $\hatbf{Y}^{j \leftarrow i}_d$) for different ego-neighbor pairs $\{i, j\}$.
To validate this design, as illustrated in \FIG{fig_exp_rehearsals}, we visualize some example short-term rehearsals forecasted by the ego predictor in several ETH-UCY and SDD scenes, under the $\{t_h, t_f\} = \{8, 12\}$ and $\{t_a, t_b\} = \{4, 4\}$ setting.
We assign different agents as the ego agent in each subfigure to directly compare differences of these biased rehearsals.
Also, results in \FIG{fig_exp_rehearsals} are forecasted with the same full \MODEL-Rev model for the direct comparisons, with the same random seed when visualizing.

\NEWHALFLINE\subsubsection{Overall Verifications of Rehearsals}

We first discuss the overall prediction quality of these rehearsals, evaluating whether the ego predictor could forecast interpretable trajectories.
As elaborated in \SECTION{sec_method_horizon}, the training of the ego predictor is only supervised by the trajectories of all agents during the global observation period $\mathcal{T}_h$.
Also, given the lightweight network structure and the only 1-order social interaction ($i \leftrightarrow j$) considered in the ego predictor, the learning process of the ego predictor could be more challenging compared to the final predictor.
However, predictions in \FIG{fig_exp_rehearsals} (b1) could still tell these minor differences from the only $\|\mathcal{T}_c\| = 4$ observation period and the limited number of ($K_I = 3$) generated trajectories.
For example, neighbors \{2, 3, 4\} have almost the same observed trajectories during the last $t_a = 4$ steps, while their predictions \{\textcolor[HTML]{c1a1d3}{$\hatbf{Y}^{2 \leftarrow 1}_d$}, \textcolor[HTML]{e3f75a}{$\hatbf{Y}^{3 \leftarrow 1}_d$}, \textcolor[HTML]{7731c6}{$\hatbf{Y}^{4 \leftarrow 1}_d$}\} present distinguishable distribution differences.

Specifically, we see that the overall distributions of these trajectories are different, where neighbor 2 has been forecasted with the widest spatial distribution of rehearsals, including one for slightly turning left, one for moving forward, and the other one for sharply turning right.
While maintaining these three trends, the distribution of predictions for neighbor 3 has slightly shrunk, representing its subtle differences compared to neighbor 2.
Notably, for neighbor 4, the ego predictor provides distinctly different predictions, where all these cases degenerate into a single moving forward mode.
Similar prediction differences can be found in almost every subfigure in \FIG{fig_exp_rehearsals}, without being restricted to any specific agent.
Note that the distance between the ego and the neighbor does not affect such predictions, since the spatial translation of trajectories is applied before inputting them to the ego predictor.
Therefore, these results indicate that the ego predictor is still capable of capturing differences in trajectories, despite the relatively limited network scales and social interactions.

\NEWHALFLINE\subsubsection{Rehearsals and Ego Biases}
\label{sec_discussion_rehearsals_and_biases}

We now discuss ego biases presented by the ego predictor for considering one specific neighbor's short-term future rehearsals from different egos' points of view.
\FIG{fig_exp_rehearsals} (a1) - (a4) visualize a four-agent prediction scene, where we take turns assigning each of these four agents as the focused ego agent.
We see that agents 1 and 4 are moving in almost exactly opposite directions, but the other two agents have fewer directional conflicts with agent 1.
Focusing on agent 4, it can be seen that its predictions $\{\hatbf{Y}^{4 \leftarrow 1}_d, \hatbf{Y}^{4 \leftarrow 2}_d, \hatbf{Y}^{4 \leftarrow 3}_d, \hatbf{Y}^{4 \leftarrow 4}_d\}$ still present notable differences, even though the motion patterns of these four agents are relatively simple and linear.

We use the \textcolor[HTML]{70d1a4}{green colorbar} running through these cases as a reference for comparison (its height matches the prediction $\hatbf{Y}^{4 \leftarrow 1}_d$ in \FIG{fig_exp_rehearsals} (a1)).
Here, we notice that agent-4's future range of motion has been considered wider from agent-1's point of view (in subfigure (a1)), even wider than agent-4's own ``thought'' (in (a4)).
In contrast, predictions $\{\hatbf{Y}^{2 \leftarrow 1}_d, \hatbf{Y}^{3 \leftarrow 1}_d\}$ in (a1), $\{\hatbf{Y}^{2 \leftarrow 2}_d, \hatbf{Y}^{3 \leftarrow 2}_d\}$ in (a2), etc., present more stable prediction distributions, even though the ego agent has been changed across these cases.
This is in line with how humans perceive their interaction context, where people may pay more attention to those who display uniquely different motions from themselves (agents $1 \leftrightarrow 4$) while ignoring those who share similar ones (agents $1 \leftrightarrow 2 \leftrightarrow 3$).
Similar predictions can be found in \FIG{fig_exp_rehearsals} (b1) to (b4) when forecasting neighbor-6's trajectories from other egos' perspectives.
Therefore, we can infer that these ego biases are not arbitrary or stochastic trajectory biases attached to the prediction (like the $j \leftarrow j$ sourced generation kernel in the original reverberation transform \REVCITE), but rather interactive assumptions jointly determined by the insight kernel $\mathbf{I}^i_\EGO$ and the original reverberation kernel $\mathbf{R}^j_\EGO$ (sourced from both $i$ and $j$, see \EQUA{eq_ego_bias_representation}), clearly related to the state differences between agents $\{i, j\}$.

\NEWHALFLINE\subsubsection{Interpretations of Biased Rehearsals}

Next, we further discuss how the predicted rehearsals reflect different ego biases for considering trajectory plans and interaction contexts, taking cases in \FIG{fig_exp_rehearsals} (c1) to (c4) as examples.
We now focus on agent 4.
It is walking in a more unstable manner (first walking to its left and then swaying to the right) than others during the 8-frame global observation period.
Regardless of who the ego is, predictions of the ego predictor for agent 4 are wider (\IE, exhibit a larger spatial spread) than other agents', which aligns with our observations in ``Rehearsals and Ego Biases''.
By comparing agent-4's predictions in \FIG{fig_exp_rehearsals} (c1) to (c4), it can be seen that relatively conservative predictions are more likely to occur when the trajectories of the ego and this agent 4 are more similar.
For example, predictions $\hatbf{Y}^{4 \leftarrow 4}_d$ in (c4) show the least divergence, and the vertical height of its dashed box is about 10\% smaller than the colorbar that serves as a reference.

Also, in these cases, we observe that the degree of prediction divergence (as indicated by the height of the dashed boxes) is nearly identical across other egos' perspectives except for agent 4 itself, though there are still notable differences in their absolute spatial distributions along the vertical axis.
These vertical biases can be naturally attributed to differences in the initial motion direction of the predicted trajectories.
By comparing the colorbar spanning from \FIG{fig_exp_rehearsals} (c1) to (c4), we observe that the predictions for agent 4 are gradually shifting downward, with their initial movement directions converging to match those of agent 4 at the current observation step, therefore maintaining a more continuous and smooth forecast trajectory.
For an ego agent whose historical trajectory (movement modes) differs more significantly from those of a particular neighbor, the ego predictor is more likely to learn and forecast the ego bias as introducing both greater discontinuity (sudden direction changes) and uncertainty (divergence) in the future movements.
We can conclude that these rehearsals not only reflect the biases of how different egos imagine their short-term future interaction contexts, but also have the capacity to differentiate these biases across different contexts.

Interestingly, such presented discontinuity and uncertainty perfectly match the temporal and spatial properties of the predicted trajectories, which correspond exactly one-to-one with the two kernels in the bilinear product in \EQUA{eq_ego_bias_representation}.
Specifically, kernel $\mathbf{R}^j_\EGO$ captures the objective temporal properties of the neighbor-$j$'s trajectory, while kernel $\mathbf{I}^i_\EGO$ acts as a moderator upon the initial product, making the predictions $\hatbf{Y}^{j \leftarrow i}_d$ exhibit distinct subjective distributions, especially spatially.

\subsection{Discussions of Ego Biases}

\begin{figure*}[tbp]
    \centering
    \includegraphics[width=1.0\linewidth]{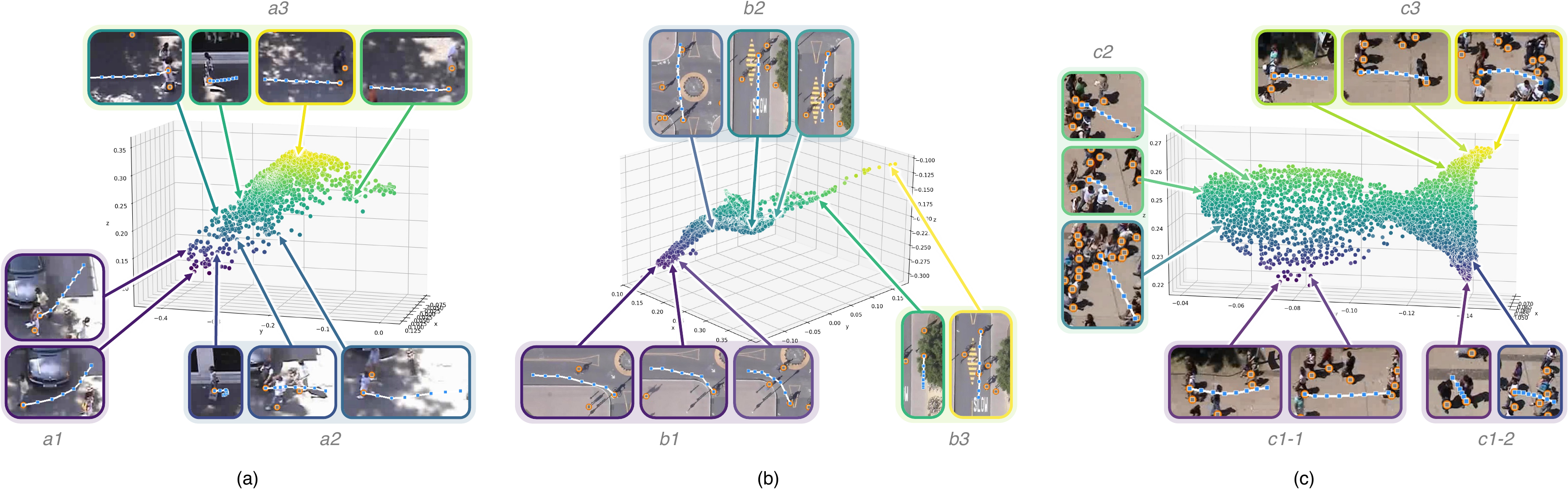}
    \caption{
        Visualized distributions of insight kernels $\mathbf{I}^i_\EGO \in \mathbb{R}^{t_a \times K_I}$ in different scenes (UCY-zara1 in (a), SDD-little0 in (b), and UCY-univ3 in (c)).
        Each kernel $\mathbf{I}^i_\EGO$ is averaged along the temporal dimension ($t_a$), where only the $K_I = 3$ different ego biases are represented as 3D points in the space.
        Colors of these 3D points denote their coordinate values along the Z-axis within these 3D spaces, where yellow indicates higher values and purple indicates lower.
    }
    \label{fig_exp_bias_distribution}
\end{figure*}

The above discussions attempt to explain ego biases and the corresponding short-term rehearsal trajectories from the agent-level, with the initial conclusion that the forecasted rehearsals are strongly associated with the differences of biases represented by different ego agents (formulated as different insight kernels $\mathbf{I}^i_\EGO$) even for the same neighbor $j$.
To further verify this correlation while exploring how these biases are related to the motion status of different ego agents, we now discuss the distributions of these biases across different datasets, identifying the properties of different egos within this distribution.
As elaborated in \EQUA{eq_ego_bias_representation} and \SECTION{sec_method_interpretations}, for each feature dimension $n$, the insight kernel ${\mathbf{I}^i_\EGO} \in \mathbb{R}^{t_a \times K_I}$ applies row operations on the objective temporal representation $\left(\left(\mathbf{F}^j_\EGO\right)_{::n} \mathbf{R}^j_\EGO \right) \in \mathbb{R}^{t_a \times t_b}$.
In other words, $t_a$ rows of temporal features (with shape $t_b \times d$) will be recombined in $K_I$ ways after the left-multiplication of the insight kernel, and our focused ego biases are correspondingly contained within these $K_I$ columns of insight kernels.
Here, to focus on the properties of these ego biases while filtering those temporal fluctuations in the relatively short observation period ($|\mathcal{T}_c| = 4$ in our experiments), we visualize the distribution of the temporal-averaged insight kernels
\begin{equation}
    \widetilde{\mathbf{I}}^i_\EGO = \frac{1}{t_a} \sum_{t = 1}^{t_a} \left(\mathbf{I}^i_\EGO\right)_{t:} \in \mathbb{R}^{K_I}
\end{equation}
in \FIG{fig_exp_bias_distribution}, therefore interpreting how these ego biases are distributed and  associated with different ego agents.

\NEWHALFLINE\subsubsection{Overall Distributions}

\FIG{fig_exp_bias_distribution} visualizes 3D distributions of these temporal-averaged insight kernels under the $K_I = 3$ setting
, where each point represents an agent's kernel as the ego agent.
Overall, we observe that these 3D points are neither randomly nor linearly distributed in the 3D space.
Rather, they consistently distribute on specific low-dimensional manifolds in the 3D space across different datasets.
Specifically, these manifolds are strongly correlated with trajectory distributions of each dataset.
For example, zara1 (\FIG{fig_exp_bias_distribution} (a)) exhibits a curved fan-shaped distribution, while little0 (subfigure (b)) is shaped like a ``twisted croissant''.
Differently, the univ (subfigure (c)) dataset, which features more complex motion patterns and interactions, presents a distinct whale-like kernel distribution.
These observations indicate that ego biases learned by the ego predictor are not an inexplicable regularization term or just a random noise vector, but rather ego priors with adaptive structures across datasets, supervised by the ego loss.

Furthermore, given that the role of insight kernels is to simultaneously perform multiple (in $K_I$ ways) biased recombinations of agents' temporal representations, these distinct low-dimensional manifold structures (rather than discrete clusters) could be further interpreted as a consequence of the restricted rank property of the bilinear multiplication\footnote{
    See more mathematical analyses in Appendix A of the previous work \REVCITE.
} in \EQUA{eq_ego_outer}.
Specifically, the short-term rehearsals are forecasted by the ego predictor through a linear decoding layer after this bilinear multiplication.
This means that the rank of these rehearsals will be naturally limited by the minimum rank among the insight kernel $\mathbf{I}^i_\EGO$, the reverberation kernel $\mathbf{R}^j_\EGO$, and the neighbor-$j$'s trajectory representation $\mathbf{F}^j_\EGO$.
Even when several higher-rank clusters appear in the distributions of insight kernels, these varied insights could still not lead to notable differences in the forecasted rehearsals, as the insight kernel $\mathbf{I}^i_\EGO$ could only play the role to recombine rows of the temporal feature $\mathbf{F}^j_\EGO$.
Therefore, alongside the optimization of the ego loss (\EQUA{eq_ego_loss}), the distribution of $\mathbf{I}^i_\EGO$ tends to manifest as a continuous state of the ``low-dimensional manifold within a high-dimensional space'', and finally reaches a dynamic equilibrium as the training progresses.

\NEWHALFLINE\subsubsection{Structures of Bias Distributions}

In addition to the overall distribution, as visualized in the boxes surrounding these distributions in \FIG{fig_exp_bias_distribution}, similar trajectory patterns\footnote{
    Only $t_a = 4$ frames of observations are used by the ego predictor.
    Full observations are still visualized for a clear and consistent presentation.
} can also be found corresponding to insight kernels in specific regions across these manifolds.
For a clear presentation, we assign different colors to these 3D distribution points based on their z-axis values in \FIG{fig_exp_bias_distribution}, and divide these distributions into three rough color regions, including the purple regions (\emph{a1}, \emph{b1}, \emph{c1-1}, \emph{c1-2}), the yellow regions (\emph{a3}, \emph{b3}, \emph{c3}), and the transitional blue-green regions (\emph{a2}, \emph{b2}, \emph{c2}).

\emph{(i) Purple Regions:}
Trajectories corresponding to the purple regions of the insight kernel distribution (\emph{a1}, \emph{b1}, as well as both \emph{c1-1} and \emph{c1-2} in \FIG{fig_exp_bias_distribution}) present more nonlinear movements than others.
We also observe that such nonlinear movements include not only intention changes like turning around in (a)-\emph{a1} and (b)-\emph{b1}, but interactive cases like avoiding the approaching pedestrian in (c)-\emph{c1-1}.
Correspondingly, ego agents with such trajectories will be predicted to have relatively \emph{boundary} ego biases, which are ultimately formalized as their unique insight kernels positioned at the margins of the distribution.
Consequently, from these ego agents' points of view, greater variability might be seen when forecasting their neighbors' short-term future rehearsal trajectories.
This further interprets the phenomenon we observe in \FIG{fig_exp_rehearsals} (c4) and discuss in \SECTION{sec_discussions_biases}, where the predicted rehearsals of agent 4 are the most different ones only when agent 4, who keeps swaying and turning throughout the whole observation period, acts as the ego agent.

\emph{(ii) Yellow Regions:}
As another margin case, trajectories corresponding to those yellow regions (\emph{a3}, \emph{b3}, and \emph{c3} in \FIG{fig_exp_bias_distribution}) display almost linear motions during the observation period of the ego predictor.
Notably, these boxed example trajectories exhibit different walking speeds, directions, and even types of agents (pedestrians in the zara1 scene, and bikers in the little0 scene).
This means that the ego predictor has the ability to learn and capture those invariant motion patterns from the relatively short ($t_a < t_h$) observation trajectories when estimating ego biases, rather than those associated with simple or direct trajectory embeddings or clusterings.
Furthermore, since we do not apply preprocessing like rotations or spatial normalizations to the trajectories input to the ego predictor, this indicates that ego biases (formulated as insight kernels) are adaptively learned as structured representations from the supervision of the proposed ego loss, capable of representing specific, stable, but adaptive ego patterns across scenes.

\emph{(iii) Blue-Green Regions:}
In \FIG{fig_exp_bias_distribution} \emph{a2}, \emph{b2}, and \emph{c2}, we also see that for the kernels that occupy most of the distribution (blue-green regions), their corresponding trajectories display stronger transitional characteristics compared to those in either the purple or the yellow regions, containing less nonlinear motions while maintaining their own unique minor differences.
For example, trajectories in \emph{a2} present a clear linear-changing trend.
Starting from the pronounced turning trajectories in the purple region, the first sampled trajectory in \emph{a2} shows a trend of turning-first then moving linearly (trajectory \#1 in \emph{a2}), then another shows an acceleration before linearly moving (\#2 in \emph{a2}), and finally the third one converges toward those in the yellow region (\emph{a3}), featuring slight turns while maintaining the overall linear motion (\#3 in \emph{a2}).

Interestingly, distributions of these transitional regions present unique connection structures in each scene.
For instance, this transition region becomes relatively compact in the zara1 scene in \FIG{fig_exp_bias_distribution} (a), which directly borders the purple and yellow regions without clear separations.
Differently, this region presents more complex manifold structures in the little0 scene in (b), where agents (pedestrians and bikers) may show more complex nonlinear motions in the traffic circle.
Notably, this region finally becomes an enormous structure (``body of the whale'') attached to other regions (``tail of the whale'') in the more complex and interactive univ3 scene in (c).
This means that ego biases manifested across different scenes (datasets) have been continuously preserved during the training process, without being reduced to several quantized representations, which further demonstrates the modeling capacity of ego biases within the insight kernel.

\NEWHALFLINE\subsection{Discussions of Bias-Conditioned Prediction}

In the \MODEL~model, we concatenate the forecasted short-term rehearsals of all agents to their original observed trajectories as the biased observations $\mathcal{X}^i_\BIASED$ to provide ego-biased trajectory and interaction conditions for the final predictor (\EQUA{eq_biased_observations}).
The above discussions provide interpretations of how ego biases (insight kernels) and forecasted rehearsals are formed and distributed.
We now discuss how distinct rehearsals lead to different bias-conditioned final predictions, by first visualizing the overall conditioning results and applying counterfactual interventions to validate causal effects among insight kernels, rehearsals, and final predictions, then exploring how such conditionings work at the feature level.

\NEWHALFLINE\subsubsection{Overall Conditionings}

The first two rows in \FIG{fig_exp_conditioning} visualize forecasted short-term rehearsal trajectories and their conditioned final \MODEL~predictions in several scenes.
Comparing \FIG{fig_exp_conditioning} (a1) and (b1), we observe that almost all predicted rehearsals directed by the ego agent in (a1) are more ``cautious'' than those by the ego in (b1), showing less divergence (multimodal characteristics) during the rehearsal period.
Correspondingly, in \FIG{fig_exp_conditioning} (a2) and (b2), such multimodal characteristics have been inherited by the final predictor, where forecasted trajectories in (b2) present either better nonlinear motions or better multimodal choices than those in (a2).
Similar trends can also be found in the turning cart case in \FIG{fig_exp_conditioning} (c1), resulting in the stronger multimodal predictions under the same number of generated trajectories.
This means that such specific ego biases, for example the ego agent might be more cautious in planning trajectories and handling interactions in \FIG{fig_exp_conditioning} (a1), have been successfully broadcasted to and finally recognized by the final predictor, therefore resulting in the differences in the final predictions.

\NEWHALFLINE\subsubsection{Counterfactual Validations}

\begin{figure}[tbp]
    \centering
    \includegraphics[width=1.0\linewidth]{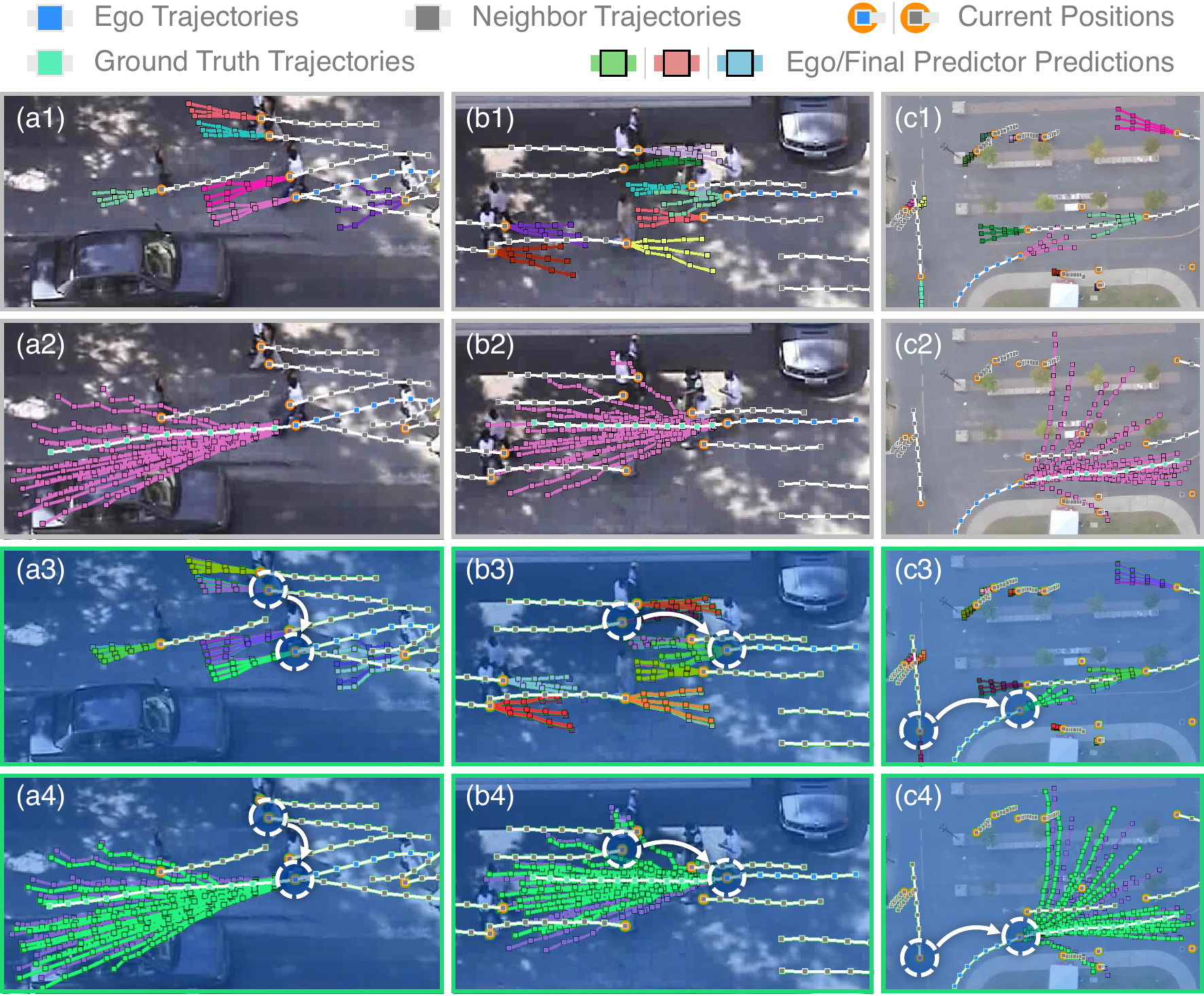}
    \caption{
        Visualizations of the bias-conditioned prediction and counterfactual interventions.
        The first two rows illustrate the original predictions made by the ego predictor and the full \MODEL~model in several scenes.
        Correspondingly, the last two rows validate the causal effect of such bias-conditioning process by applying the intervention of switching ego agents' insight kernels with other agents' (marked with white circles).
    }
    \label{fig_exp_conditioning}
\end{figure}

We conduct counterfactual validations and visualize their results in the last two rows in \FIG{fig_exp_conditioning} to further verify whether the \MODEL~learns causal relations during the above discussed bias ``broadcast'' or just simple data fitting, \IE, during the bias propagation from the insight kernel $\mathbf{I}^i_\EGO$ to the short-term rehearsals $\left\{\hatbf{Y}^{j \leftarrow i}_d\right\}$ and the final \MODEL~predictions $\hatbf{Y}^{i \leftarrow i}_\FINAL$ for the specific ego $i$ and all agents $\forall j$ in the scene.
For a clear presentation, we use $I_\EGO^i$, $Y_d^i$, and $Y_\FINAL^i$ to indicate these causal variables, respectively.
We use two kinds of interventions to run individual validations of causal edges $I_\EGO^i \to Y_d^i$ and $Y_d^i \to Y_\FINAL^i$ separately, including the kernel intervention and the rehearsal intervention.

\emph{(i) Kernel Interventions:}
Based on our initial observations in \SECTION{sec_discussion_rehearsals_and_biases}, here we first intervene the insight kernel $\mathbf{I}^i_\EGO$ directly in \EQUA{eq_ego_bias_representation}, by replacing it with the kernel $\mathbf{I}^m_\EGO$ of another specific neighbor $m$ in the scene ($m \neq i$), whose kernel presents a distinct difference from that of ego $i$ (by comparing their Euclidean distance).
Formally, denote all other used external features and matrices as variable $F_\EGO^i$, we have $\left\{\hatbf{Y}^{j \leftarrow i}_d\right\} = Y_d^i \left(I_\EGO^i = \mathbf{I}^i_\EGO, F_\EGO^i\right)$ before the intervention.
Correspondingly, this kernel intervention and its counterfactual output can be formulated as $\left\{\overline{\hatbf{Y}^{j \leftarrow i}_d}\right\} = Y_d^i \left(do\left(I_\EGO^i = \mathbf{I}^m_\EGO\right), F_\EGO^i\right)$.
As illustrated in \FIG{fig_exp_conditioning} (a3), (b3), (c3), we apply such kernel interventions (selected neighbor $m$ has been marked with white circles) and visualize the intervened rehearsals correspondingly.
We see that the forecasted rehearsals have been greatly modified after interventions, where rehearsals in both \FIG{fig_exp_conditioning} (a3) and (b3) present less diverse motion trends than those original predictions in (a1) and (b1), while rehearsals in case (c3) show greater tendencies to maintain motion states rather than the more notable direction changes in (c1), which aligns with the behavioral differences between the selected neighbor $m$ and the ego $i$.
Note that in these interventions, we have strictly maintained consistency across all other variables ($F_\EGO^i$) and used the identical fixed random seed when computing.
Therefore, like previous works \cite{chen2021human,wong2024socialcircle+}, we can conclude that the ego predictor has learned the causal mechanism along the pathway $I^i_\EGO \to Y^i_d$, according to the differences between the original rehearsals and these intervened ones.

\emph{(ii) Rehearsal Interventions:}
The rehearsal interventions aim to verify whether causal effect $Y^i_d \to Y^i_\FINAL$ has been learned by the final predictor.
Here, we do not introduce extra intervention methods to directly modify rehearsals manually, but rather use the above kernel-intervened rehearsals $\left\{\overline{\hatbf{Y}^{j \leftarrow i}_d}\right\}$ to verify the causal chain $I_\EGO^i \to Y_d^i \to Y_\FINAL^i$ continuously.
For the final predictor, its original prediction can be formulated as $\hatbf{Y}^{i \leftarrow i}_\FINAL = Y_\FINAL^i \left(Y^i_d = \left\{\hatbf{Y}^{j \leftarrow i}_d\right\}, F^i_\FINAL\right)$, where variable $F^i_\FINAL$ denotes all other model inputs (which will also be fixed before and after the intervention).
Similar to the above kernel intervention, the rehearsal intervention and its output has become $\overline{\hatbf{Y}^{i \leftarrow i}_\FINAL} = Y_\FINAL^i \left(do\left(Y^i_d = \left\{\overline{\hatbf{Y}^{j \leftarrow i}_d}\right\}\right), F^i_\FINAL\right)$.
Corresponding to the rehearsal changes from \FIG{fig_exp_conditioning} (a1) to (a3), (b1) to (b3), and (c1) to (c3), distinct prediction modifications can be observed from \FIG{fig_exp_conditioning} (a4), (b4), and (c4).
For example, predictions in case (b4) present a considerable decrease of multimodality to follow up its intervened rehearsals (illustrated in (b3)).
The multipath properties of predictions in \FIG{fig_exp_conditioning} (c4) also present unique differences corresponding to its intervened rehearsals in (c3), where three or four clear trajectory clusters are more visibly presented rather than being relatively uniformly distributed across the space in \FIG{fig_exp_conditioning} (c2).
Hence, causal edge $Y_d^i \to Y_\FINAL^i$ could be verified, which further means that a consistent causal dependency chain $I_\EGO^i \to Y_d^i \to Y_\FINAL^i$ has been learned.
Therefore, \MODEL~has the ability to not only represent but also broadcast those ego biases throughout the whole network rather than the gradual degeneration.

\NEWHALFLINE\subsubsection{Rehearsal Selections}

From the above discussion we observed that distinct biased observations $\mathcal{X}^i_\BIASED$ could lead to different \MODEL~predictions, even for the same ego-neighbor pair.
Here, for either the ego $i$ or any of its neighbors $j$, such biased observations are made up of $K_I$ full rehearsal trajectories $\left(\hatbf{X}^{j \leftarrow i}_r\right)_{k::}$.
These full rehearsal trajectories are fused via the feature-level max pooling (\EQUA{eq_method_maxpool}), and then used as queries and keys in Transformer attention layers (within the final predictor) to condition final predictions, taking their mean trajectories as spatial references (\EQUA{eq_method_mean}).
We now discuss how different full rehearsal trajectories contribute to constructing the final conditioned representation ($\mathbf{f}^{j \leftarrow i}_\FINAL$ in \EQUA{eq_method_condition}) at the feature level.

\begin{figure*}[tbp]
    \centering
    \includegraphics[width=1.0\linewidth]{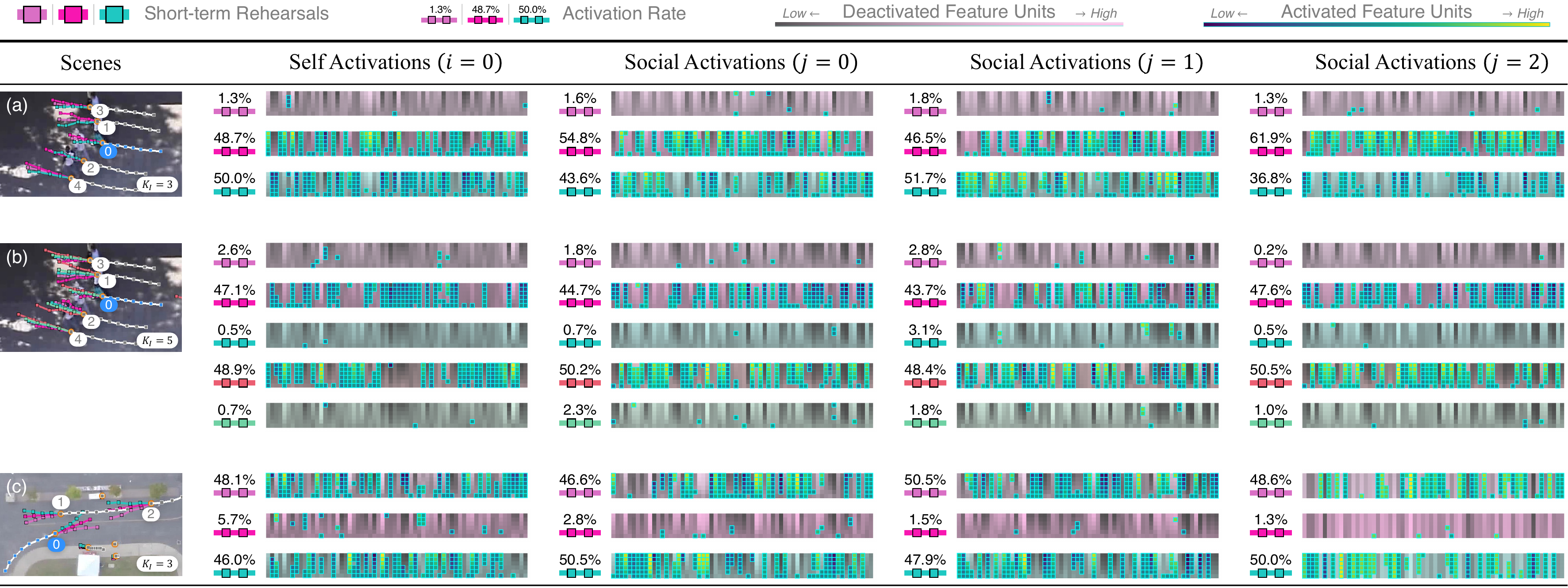}
    \caption{
        Visualized feature activations and their corresponding activation rates $r^{j \leftarrow i}_k~(k = 1, 2, ..., K_I)$ during the feature-level bias conditioning.
        Each matrix represents the $k$th representation $\left(\mathbf{f}^{j \leftarrow i}_r\right)_{k::}$ of the $k$th full rehearsal trajectory $\left(\hatbf{X}^{j \leftarrow i}_r\right)_{k::}$.
        Agent 0 is treated as the ego agent in all these scenes.
    }
    \label{fig_exp_activations}
\end{figure*}

As illustrated in \FIG{fig_exp_activations}, to directly view how each full rehearsal contributes to the final predicted trajectories, we visualize how the activated feature units are located after the max pooling and use a simple metric, the \emph{Activation Rate}, to measure how many units have been selected in each feature.
In addition, \FIG{fig_exp_activations} visualizes not only activation rates of egos' ($i = 0$) own trajectories but their neighbors' ($j = 0, 1, 2$), which explicitly model the egos' self-intention changes and their interactions with neighbors, respectively.
In detail, for the representation of the $k$th full rehearsal trajectory $\left(\mathbf{f}^{j \leftarrow i}_r\right)_{k::} \in \mathbb{R}^{(t_h + t_b) \times d'}$, its activation rate is computed as
\begin{equation}
    r^{j \leftarrow i}_k = \frac{1}{N_r} \sum_m \sum_n \mathbb{I}\left(k, \left(\mathbf{f}^{j \leftarrow i}_r\right)_{:, m, n}\right),
\end{equation}
where $\mathbb{I}$ denotes an indicator function
\begin{equation}
    \mathbb{I}\left(k, \left(\mathbf{f}^{j \leftarrow i}_r\right)_{:, m, n}\right) = \left\{\begin{aligned}
        &1,\quad k = \operatorname*{arg\,max}_{k'=1}^{K_I} \left(\mathbf{f}^{j \leftarrow i}_r\right)_{k', m, n}, \\
        &0,\quad \mbox{Otherwise},
    \end{aligned}\right.
\end{equation}
and $N_r = d' (t_h + t_b)$ denotes the total number of feature units.

In \FIG{fig_exp_activations}, we find that the activation rate of a specific full rehearsal trajectory is strongly associated with the differences between the motion states within the forecasted short-term rehearsals and the original observed trajectories of an agent.
For example, for ego 0 in \FIG{fig_exp_activations} (a), in the $K_I = 3$ setting, the ego predictor forecasts it with three distinct short-term rehearsals, including one for slightly faster turning right, one for slightly slower turning left, while a third for keeping the linear motion.
Although the absolute spatial differences between these rehearsals remain relatively small, for example the distance between the endpoints of the right and left turn rehearsals is less than 0.5 meters, the features of these rehearsals exhibit distinctly different trends.
Under the max pooling feature-level conditioning, the right-turn one gets about 48.7\% activated feature units for consideration by the final predictor, while the left-turn one gets a comparable 50.0\%, presenting a strong complementary feature-selection trend.
In other words, unlike methods that directly concatenate interaction features or even apply the average pooling on features, this feature-level max pooling allows for the simultaneous and direct capture of effective patterns from both the left and right turning rehearsals, rather than compressing or smoothing them out.
In particular, the total activations of these two rehearsals reached nearly 99\%, indicating that the rehearsal of linear movement has been virtually ignored (only 1.3\%).

Interestingly, not all linear movements will be ignored.
For the ego 0 in \FIG{fig_exp_activations} (c), we see that its bottom linear-movement rehearsal obtains about 48.1\% activations, while the middle one (which also presents similar linear motion with a different moving direction) only gets 5.7\%.
Here, the significant difference between the motion states of egos in cases (a) and (c) is that ego 0 in (a) maintains the almost linear motion during observation, while the other ego in (c) keeps turning on the curved road.
This is in line with our intuition as pedestrians that when planning our future, we tend to focus on those who exhibit significant behavior shifts, whether in the real world or just in our imaginations, rather than those who maintain consistent modes.
We can conclude from these observations that the feature-level max pooling allows the final predictor to adaptively adjust its focus for different short-term rehearsals forecasted by the ego predictor relative to the original motion states of a specific agent, rather than providing equal or single-mode treatments.

\NEWHALFLINE\subsubsection{Feature Activations and Social Interactions}

\FIG{fig_exp_activations} also indicates that the same $k$th rehearsal may get different feature activation rates, depending on the purpose as well as the overall states of all the rehearsals.
For example, rehearsals for egos themselves ($\mathbf{f}^{0 \leftarrow 0}_r$, the $i = 0$ column) may be used to estimate their own future plans, while rehearsals for their neighbors ($\mathbf{f}^{j \leftarrow 0}_r$, including themselves, the $j = 0, 1, 2$ columns) are more likely to consider potential social interactions.
Especially, for the employed final predictor \REVMODEL~\REVCITE, such self-intention changes and all other interactive behaviors are separately modeled and forecasted, providing convenient bases for analyzing bias-conditionings with distinct trajectory goals.

In \FIG{fig_exp_activations} (a), among various rehearsals directed by ego 0 toward other agents, those rehearsals pursued for different purposes have been assigned significantly different activation rates, particularly given that these rehearsals are all predicted using exactly the same ego predictor and the same ego bias (\IE, insight kernel $\mathbf{I}^0_\EGO$).
For example, when considering social interactions, ego 0 may consider the rehearsal of turning right to be more important (54.8\%), with an 11.2\% higher activation rate than the other downward left-turn rehearsal (43.6\%), rather than sharing almost the same focus when only considering its own intention changes in the $i = 0$ column.
We infer that this may be because ego 0 is closer to neighbors 1 and 3 above it.
Under such situations, when the final predictor evaluates social interactions internally, it pays more attention to this region rather than to other agents far away below (this is determined by the \REVMODEL~model as the final predictor, which employs angle-based distance measurements and social modeling networks).

Simultaneously, when rehearsing neighbor $j = 1$ from ego-0's point of view in case (a), its turning down (left) rehearsal (which is closer to ego 0 itself, carrying a greater potential for collision) gets 51.7\% activation, which is 5.2\% larger than the further side relative to the ego 0.
Similar phenomena can also be found for the neighbor $j = 2$, whose turning right rehearsal gets a notable 61.9\% activation rate.
This means that the final predictor has the flexibility to adaptively evaluate the contribution of each rehearsal under the feature-level conditioning, depending on whether the self-intended or socially interactive trajectories are required to be forecasted, beyond the only differences in motion patterns between rehearsals and the original observations.
In other words, \MODEL~has the ability to perform collaborative optimization alongside this feature selection, which serves as a \emph{bridge} between the ego predictor and the final predictor, rather than isolated training or hard data fusion and sharing, particularly integrating with the interaction modeling component within the final predictor.

\NEWHALFLINE\subsubsection{Activation Rates, Max Pooling, and $K_I$}
\label{sec_discussion_maxpool}

\begin{table}[tbp]
    \caption{
        Activation Rate Comparisons (Mean Rehearsals)
    }
    \label{tab_mean_activations}
    \centering
    \begin{threeparttable}
        \setlength{\tabcolsep}{4pt}
        \begin{tabular}{ccccccccccccccc}
            \toprule
            ID & $i = 0$ & $j = 0$ & $j = 1$ & $j = 2$ \\

            \midrule
            a-1 & 0.010 (\X{-0.003}) & 0.010 (\X{-0.006}) & 0.009 (\X{-0.009}) & 0.007 (\X{-0.006}) \\
            a-2 & 0.487 (+0.000) & 0.545 (\X{-0.003}) & 0.462 (\X{-0.003}) & 0.619 (+0.000) \\
            a-3 & 0.500 (+0.000) & 0.436 (+0.000) & 0.514 (\X{-0.003}) & 0.365 (\X{-0.003}) \\
            mean & 0.003 (\C{+0.003}) & 0.009 (\C{+0.009}) & 0.015 (\C{+0.015}) & 0.009 (\C{+0.009}) \\
            
            \midrule
            b-1 & 0.014 (\X{-0.012}) & 0.009 (\X{-0.009}) & 0.022 (\X{-0.006}) & 0.002 (+0.000) \\
            b-2 & 0.468 (\X{-0.003}) & 0.435 (\X{-0.012}) & 0.434 (\X{-0.003}) & 0.470 (\X{-0.006})  \\
            b-3 & 0.005 (+0.000) & 0.007 (+0.000) & 0.028 (\X{-0.003}) & 0.005 (+0.000) \\
            b-4 & 0.486 (\X{-0.003}) & 0.487 (\X{-0.015}) & 0.475 (\X{-0.009}) & 0.502 (\X{-0.003})  \\
            b-5 & 0.007 (+0.000) & 0.020 (\X{-0.003}) & 0.015 (\X{-0.003}) & 0.010 (+0.000) \\
            mean & 0.020 (\C{+0.020}) & 0.042 (\C{+0.042}) & 0.026 (\C{+0.026}) & 0.011 (\C{+0.011})  \\

            \midrule
            c-1 & 0.475 (\X{-0.006}) & 0.463 (\X{-0.003}) & 0.502 (\X{-0.003}) & 0.486 (+0.000) \\
            c-2 & 0.034 (\X{-0.023}) & 0.016 (\X{-0.012}) & 0.009 (\X{-0.006}) & 0.007 (\X{-0.006}) \\
            c-3 & 0.457 (\X{-0.003}) & 0.505 (+0.000) & 0.476 (\X{-0.003}) & 0.500 (+0.000) \\
            mean & 0.034 (\C{+0.034}) & 0.016 (\C{+0.016}) & 0.013 (\C{+0.013}) & 0.007 (\C{+0.007}) \\
                
            \bottomrule
        \end{tabular}

        \begin{tablenotes}[flushleft]
            \footnotesize
            \item[]
                Activation rates are computed after adding features of mean rehearsal trajectories to the candidate features for max pooling.
                Numbers in parentheses indicate the changes relative to the original activations in \FIG{fig_exp_activations}.
        \end{tablenotes}
    \end{threeparttable}
\end{table}

The above discussions indicate that the feature-level max pooling actually serves as an adaptive feature gate, filtering the \emph{notable} ones from all $K_I$ rehearsals for a specific agent.
We now further interpret why we choose max pooling.
Like the social pooling in \SOCIALLSTM~\SOCIALLSTMCITE~and many of its variations, pooling features in specific regions has become a common choice especially for fusing representations of different agents like social interaction features.
Among these methods, a considerable number of researchers select the mean (average) pooling.
However, in our bias-conditioning framework, this averaging may lead to the flattening of the ego biases presented in insight kernels constructed by the ego predictor, whether averaging the rehearsals themselves or their representations.
In the proposed \MODEL, the ego predictor first forecasts $K_I$ distinct biased short-term rehearsals for each agent, emulating the ego agents' points of view.
As mentioned in our previous discussions, these $K_I$ rehearsals exhibit significant divergence, whether in the distribution of insight kernels in \FIG{fig_exp_bias_distribution} or in the predicted short-term rehearsals in \FIG{fig_exp_conditioning}, where such differences could be easily eliminated by averaging.

To further validate this assumption, we report the activation rates after adding the features of mean rehearsal trajectories to the candidate features for max pooling in \TABLE{tab_mean_activations}.
Compared to the original activation rates reported in \FIG{fig_exp_activations}, it can be seen that the mean rehearsals have almost no or very few activated feature units, while the activation rates of those \emph{notable} rehearsals remain consistent (less than 1\% activation rate changes).
This also validates that the final predictor selects the features that are worth paying attention to, rather than focusing solely on the overall or average features of all rehearsals, which weakens the specific properties present in those rehearsals.
Furthermore, we find that the activations of these mean rehearsals exhibit extremely similar feature selections to the linear motion rehearsals (like the first row in \FIG{fig_exp_activations} (a) and the second row in (c)), sharing nearly identical bias-conditioning roles.
It means that these mean rehearsals exhibit the uniform, continuous motion trends like the forecasted linear motion ones, regardless of whether $K_I$ is set to 3 or 5, and regardless of the differences in prediction scenes.
Because of these stable, consistent, and non-activated properties, these mean rehearsals are naturally suited to serve as references for evaluating other distinct forecasted rehearsals.
Therefore, these results further validate our choice of conditioning the final predictor via biased rehearsals in two different aspects, the feature-level max pooling to aggregate competitive and noteworthy ego biases from all the forecasted rehearsals, as well as the trajectory-level average (\EQUA{eq_method_mean}) as a stable reference baseline for anchoring these activated features.

We also observe an interesting phenomenon in both \FIG{fig_exp_activations} and \TABLE{tab_mean_activations} that only two main rehearsals may get most of the activations, regardless of how the number of forecasted rehearsals ($K_I$) varies.
In other words, regardless of how many rehearsals are forecasted for the same agent, only two rehearsals could actually work to condition the final predictor.
In addition, as shown in \FIG{fig_exp_activations}, these two rehearsals with the highest activation rates are those \emph{boundary} ones that could almost cover all other rehearsals.
This is consistent with our intuition that we tend to assign greater attention to those rehearsals that exhibit significant differences in their motion states, particularly those in more extreme conditions.
In other words, these boundary rehearsals and the rehearsal of linear motion can be considered as a set of basis vectors to combine all the rehearsal trajectories, where the boundary ones always get the largest combination coefficients.
This is consistent with the results in ablation studies (\SECTION{sec_ablation_K}) that a larger $K_I$ does not necessarily yield better results, where $K_I = 3$ has become the best choice for considering ego biases.

\subsection{Conclusions}

In this manuscript, we interpret trajectory prediction into the continuous \emph{rehearsal to encore} phases through the psychological prefactual thoughts and theatrical rehearsal concepts.
The proposed \MODEL~trajectory prediction model contains a novel ego predictor, which explicitly formulates ego-agents' subjectivities as the direct set of biased short-term rehearsal trajectories directed by a specific ego agent, through the asymmetric bilinear mapping that gathers information from both ego-$i$'s biases for considering interactions (the insight kernel $\mathbf{I}^i$) and any observed neighbor-$j$'s objective temporal representations (the reverberation kernel $\mathbf{R}^j$).
The final \MODEL~predictions are conditioned by these forecasted rehearsals, from both the direct trajectory and the further feature levels, therefore taking into account specific ego biases when forecasting trajectories, achieving the goal of interpreting agents' structured and anisotropic subjectivities.
Experiments and discussions indicate that \MODEL~not only achieves consistent quantitative improvements across different multi-agent datasets, but more importantly, provides considerable interpretability regarding agents' subjectivities, which are gradually formulated as specific ego biases with specific manifolds, as the corresponding biased asymmetric yet structured rehearsal trajectories, and finally as the biased causal predictions.
In future work, we plan to extend this rehearsal-encore framework into broader interactive tasks or domains, where deciphering the subjectivity of diverse participants is critical for advancing cognitive and interpretable machine intelligence.

\bibliographystyle{IEEEtran}
\bibliography{ref.bib}

\begin{thebibliography}{10}
\providecommand{\url}[1]{#1}
\csname url@samestyle\endcsname
\providecommand{\newblock}{\relax}
\providecommand{\bibinfo}[2]{#2}
\providecommand{\BIBentrySTDinterwordspacing}{\spaceskip=0pt\relax}
\providecommand{\BIBentryALTinterwordstretchfactor}{4}
\providecommand{\BIBentryALTinterwordspacing}{\spaceskip=\fontdimen2\font plus
\BIBentryALTinterwordstretchfactor\fontdimen3\font minus \fontdimen4\font\relax}
\providecommand{\BIBforeignlanguage}[2]{{%
\expandafter\ifx\csname l@#1\endcsname\relax
\typeout{** WARNING: IEEEtran.bst: No hyphenation pattern has been}%
\typeout{** loaded for the language `#1'. Using the pattern for}%
\typeout{** the default language instead.}%
\else
\language=\csname l@#1\endcsname
\fi
#2}}
\providecommand{\BIBdecl}{\relax}
\BIBdecl

\bibitem{alahi2016social}
A.~Alahi, K.~Goel, V.~Ramanathan, A.~Robicquet, L.~Fei-Fei, and S.~Savarese, ``Social lstm: Human trajectory prediction in crowded spaces,'' in \emph{Proceedings of the IEEE conference on computer vision and pattern recognition}, 2016, pp. 961--971.

\bibitem{chen2025dstigcn}
W.~Chen, H.~Sang, J.~Wang, and Z.~Zhao, ``Dstigcn: Deformable spatial-temporal interaction graph convolution network for pedestrian trajectory prediction,'' \emph{IEEE Transactions on Intelligent Transportation Systems}, 2025.

\bibitem{zhou2024trajpred}
C.~Zhou, G.~AlRegib, A.~Parchami, and K.~Singh, ``Trajpred: Trajectory prediction with region-based relation learning,'' \emph{IEEE Transactions on Intelligent Transportation Systems}, 2024.

\bibitem{kothari2023safety}
P.~Kothari and A.~Alahi, ``Safety-compliant generative adversarial networks for human trajectory forecasting,'' \emph{IEEE Transactions on Intelligent Transportation Systems}, vol.~24, no.~4, pp. 4251--4261, 2023.

\bibitem{rempe2023trace}
D.~Rempe, Z.~Luo, X.~B. Peng, Y.~Yuan, K.~Kitani, K.~Kreis, S.~Fidler, and O.~Litany, ``Trace and pace: Controllable pedestrian animation via guided trajectory diffusion,'' \emph{arXiv preprint arXiv:2304.01893}, 2023.

\bibitem{yue2024human}
J.~Yue, B.~Li, J.~Pettr{\'e}, A.~Seyfried, and H.~Wang, ``Human motion prediction under unexpected perturbation,'' in \emph{Proceedings of the IEEE/CVF Conference on Computer Vision and Pattern Recognition}, 2024, pp. 1501--1511.

\bibitem{bae2025continuous}
I.~Bae, J.~Lee, and H.-G. Jeon, ``Continuous locomotive crowd behavior generation,'' in \emph{Proceedings of the Computer Vision and Pattern Recognition Conference}, 2025, pp. 22\,416--22\,431.

\bibitem{yagi2018future}
T.~Yagi, K.~Mangalam, R.~Yonetani, and Y.~Sato, ``Future person localization in first-person videos,'' in \emph{Proceedings of the IEEE Conference on Computer Vision and Pattern Recognition}, 2018, pp. 7593--7602.

\bibitem{patel2025uniegomotion}
C.~Patel, H.~Nakamura, Y.~Kyuragi, K.~Kozuka, J.~C. Niebles, and E.~Adeli, ``Uniegomotion: A unified model for egocentric motion reconstruction, forecasting, and generation,'' in \emph{Proceedings of the IEEE/CVF International Conference on Computer Vision}, 2025, pp. 10\,318--10\,329.

\bibitem{li2025unified}
R.~Li, T.~Qiao, S.~Katsigiannis, Z.~Zhu, and H.~P. Shum, ``Unified spatial-temporal edge-enhanced graph networks for pedestrian trajectory prediction,'' \emph{IEEE Transactions on Circuits and Systems for Video Technology}, 2025.

\bibitem{li2022graph}
L.~Li, M.~Pagnucco, and Y.~Song, ``Graph-based spatial transformer with memory replay for multi-future pedestrian trajectory prediction,'' in \emph{Proceedings of the IEEE/CVF Conference on Computer Vision and Pattern Recognition}, 2022, pp. 2231--2241.

\bibitem{xu2022groupnet}
C.~Xu, M.~Li, Z.~Ni, Y.~Zhang, and S.~Chen, ``Groupnet: Multiscale hypergraph neural networks for trajectory prediction with relational reasoning,'' in \emph{Proceedings of the IEEE/CVF Conference on Computer Vision and Pattern Recognition (CVPR)}, June 2022, pp. 6498--6507.

\bibitem{bae2022learning}
I.~Bae, J.-H. Park, and H.-G. Jeon, ``Learning pedestrian group representations for multi-modal trajectory prediction,'' in \emph{European Conference on Computer Vision}.\hskip 1em plus 0.5em minus 0.4em\relax Springer, 2022, pp. 270--289.

\bibitem{zou2024who}
Z.~Zou, C.~Wong, B.~Xia, and X.~You, ``Who walks with you matters: Perceiving social interactions with groups for pedestrian trajectory prediction,'' in \emph{Proceedings of the IEEE/CVF International Conference on Computer Vision}, 2025, pp. 4844--4853.

\bibitem{rossi2021human}
\BIBentryALTinterwordspacing
L.~Rossi, M.~Paolanti, R.~Pierdicca, and E.~Frontoni, ``Human trajectory prediction and generation using lstm models and gans,'' \emph{Pattern Recognition}, vol. 120, p. 108136, 2021. [Online]. Available: \url{https://www.sciencedirect.com/science/article/pii/S003132032100323X}
\BIBentrySTDinterwordspacing

\bibitem{gupta2018social}
A.~Gupta, J.~Johnson, L.~Fei-Fei, S.~Savarese, and A.~Alahi, ``Social gan: Socially acceptable trajectories with generative adversarial networks,'' in \emph{Proceedings of the IEEE Conference on Computer Vision and Pattern Recognition}, 2018, pp. 2255--2264.

\bibitem{lee2022muse}
M.~Lee, S.~S. Sohn, S.~Moon, S.~Yoon, M.~Kapadia, and V.~Pavlovic, ``Muse-vae: Multi-scale vae for environment-aware long term trajectory prediction,'' in \emph{Proceedings of the IEEE/CVF Conference on Computer Vision and Pattern Recognition}, 2022, pp. 2221--2230.

\bibitem{xu2022socialvae}
P.~Xu, J.-B. Hayet, and I.~Karamouzas, ``Socialvae: Human trajectory prediction using timewise latents,'' in \emph{European Conference on Computer Vision}, 2022, pp. 511--528.

\bibitem{bae2024singulartrajectory}
I.~Bae, Y.-J. Park, and H.-G. Jeon, ``Singulartrajectory: Universal trajectory predictor using diffusion model,'' \emph{arXiv preprint arXiv:2403.18452}, 2024.

\bibitem{li2024bcdiff}
R.~Li, C.~Li, D.~Ren, G.~Chen, Y.~Yuan, and G.~Wang, ``Bcdiff: Bidirectional consistent diffusion for instantaneous trajectory prediction,'' \emph{Advances in Neural Information Processing Systems}, vol.~36, 2024.

\bibitem{mangalam2020s}
K.~Mangalam, Y.~An, H.~Girase, and J.~Malik, ``From goals, waypoints \& paths to long term human trajectory forecasting,'' in \emph{Proceedings of the IEEE/CVF International Conference on Computer Vision}, 2021, pp. 15\,233--15\,242.

\bibitem{wong2023socialcircle}
C.~Wong, B.~Xia, Z.~Zou, Y.~Wang, and X.~You, ``Socialcircle: Learning the angle-based social interaction representation for pedestrian trajectory prediction,'' in \emph{Proceedings of the IEEE/CVF Conference on Computer Vision and Pattern Recognition}, 2024, pp. 19\,005--19\,015.

\bibitem{wong2024socialcircle+}
C.~Wong, B.~Xia, Z.~Zou, and X.~You, ``Socialcircle+: Learning the angle-based conditioned interaction representation for pedestrian trajectory prediction,'' \emph{arXiv preprint arXiv:2409.14984}, 2024.

\bibitem{epstude2016prefactual}
K.~Epstude, A.~Scholl, and N.~J. Roese, ``Prefactual thoughts: Mental simulations about what might happen,'' \emph{Review of general psychology}, vol.~20, no.~1, pp. 48--56, 2016.

\bibitem{helbing1995social}
D.~Helbing and P.~Molnar, ``Social force model for pedestrian dynamics,'' \emph{Physical review E}, vol.~51, no.~5, p. 4282, 1995.

\bibitem{zhang2020social}
P.~Zhang, J.~Xue, P.~Zhang, N.~Zheng, and W.~Ouyang, ``Social-aware pedestrian trajectory prediction via states refinement lstm,'' \emph{IEEE transactions on pattern analysis and machine intelligence}, vol.~44, no.~5, pp. 2742--2759, 2022.

\bibitem{huang2021lstm}
\BIBentryALTinterwordspacing
Z.~Huang, J.~Wang, L.~Pi, X.~Song, and L.~Yang, ``Lstm based trajectory prediction model for cyclist utilizing multiple interactions with environment,'' \emph{Pattern Recognition}, vol. 112, p. 107800, 2021. [Online]. Available: \url{https://www.sciencedirect.com/science/article/pii/S0031320320306038}
\BIBentrySTDinterwordspacing

\bibitem{zhang2019sr}
P.~Zhang, W.~Ouyang, P.~Zhang, J.~Xue, and N.~Zheng, ``Sr-lstm: State refinement for lstm towards pedestrian trajectory prediction,'' in \emph{Proceedings of the IEEE Conference on Computer Vision and Pattern Recognition}, 2019, pp. 12\,085--12\,094.

\bibitem{bisagno2018group}
N.~Bisagno, B.~Zhang, and N.~Conci, ``Group lstm: Group trajectory prediction in crowded scenarios,'' in \emph{Proceedings of the European conference on computer vision (ECCV) workshops}, 2018, pp. 0--0.

\bibitem{xue2018ss}
H.~Xue, D.~Q. Huynh, and M.~Reynolds, ``Ss-lstm: A hierarchical lstm model for pedestrian trajectory prediction,'' in \emph{2018 IEEE Winter Conference on Applications of Computer Vision (WACV)}.\hskip 1em plus 0.5em minus 0.4em\relax IEEE, 2018, pp. 1186--1194.

\bibitem{kipf2016semi}
T.~N. Kipf and M.~Welling, ``Semi-supervised classification with graph convolutional networks,'' \emph{arXiv preprint arXiv:1609.02907}, 2016.

\bibitem{shi2021sgcn}
L.~Shi, L.~Wang, C.~Long, S.~Zhou, M.~Zhou, Z.~Niu, and G.~Hua, ``Sgcn: Sparse graph convolution network for pedestrian trajectory prediction,'' in \emph{Proceedings of the IEEE/CVF Conference on Computer Vision and Pattern Recognition}, 2021, pp. 8994--9003.

\bibitem{lv2023skgacn}
K.~Lv and L.~Yuan, ``Skgacn: social knowledge-guided graph attention convolutional network for human trajectory prediction,'' \emph{IEEE Transactions on Instrumentation and Measurement}, 2023.

\bibitem{liu2021avgcn}
C.~Liu, Y.~Chen, M.~Liu, and B.~E. Shi, ``Avgcn: Trajectory prediction using graph convolutional networks guided by human attention,'' in \emph{2021 IEEE International Conference on Robotics and Automation (ICRA)}.\hskip 1em plus 0.5em minus 0.4em\relax IEEE, 2021, pp. 14\,234--14\,240.

\bibitem{kosaraju2019social}
V.~Kosaraju, A.~Sadeghian, R.~Mart{\'\i}n-Mart{\'\i}n, I.~Reid, H.~Rezatofighi, and S.~Savarese, ``Social-bigat: Multimodal trajectory forecasting using bicycle-gan and graph attention networks,'' in \emph{Advances in Neural Information Processing Systems}, 2019, pp. 137--146.

\bibitem{mohamed2020social}
A.~Mohamed, K.~Qian, M.~Elhoseiny, and C.~Claudel, ``Social-stgcnn: A social spatio-temporal graph convolutional neural network for human trajectory prediction,'' in \emph{Proceedings of the IEEE/CVF Conference on Computer Vision and Pattern Recognition}, 2020, pp. 14\,424--14\,432.

\bibitem{huang2019stgat}
Y.~Huang, H.~Bi, Z.~Li, T.~Mao, and Z.~Wang, ``Stgat: Modeling spatial-temporal interactions for human trajectory prediction,'' in \emph{Proceedings of the IEEE International Conference on Computer Vision}, 2019, pp. 6272--6281.

\bibitem{cao2021spectral}
D.~Cao, J.~Li, H.~Ma, and M.~Tomizuka, ``Spectral temporal graph neural network for trajectory prediction,'' in \emph{2021 IEEE International Conference on Robotics and Automation (ICRA)}.\hskip 1em plus 0.5em minus 0.4em\relax IEEE, 2021, pp. 1839--1845.

\bibitem{cao2020spectral}
D.~Cao, Y.~Wang, J.~Duan, C.~Zhang, X.~Zhu, C.~Huang, Y.~Tong, B.~Xu, J.~Bai, J.~Tong \emph{et~al.}, ``Spectral temporal graph neural network for multivariate time-series forecasting,'' \emph{Advances in Neural Information Processing Systems}, vol.~33, pp. 17\,766--17\,778, 2020.

\bibitem{chib2024lg}
P.~S. Chib and P.~Singh, ``Lg-traj: Llm guided pedestrian trajectory prediction,'' \emph{arXiv preprint arXiv:2403.08032}, 2024.

\bibitem{bae2025social}
I.~Bae, J.~Lee, and H.-G. Jeon, ``Social reasoning-aware trajectory prediction via multimodal language model,'' \emph{IEEE Transactions on Pattern Analysis and Machine Intelligence}, 2025.

\bibitem{sadeghian2019sophie}
A.~Sadeghian, V.~Kosaraju, A.~Sadeghian, N.~Hirose, H.~Rezatofighi, and S.~Savarese, ``Sophie: An attentive gan for predicting paths compliant to social and physical constraints,'' in \emph{Proceedings of the IEEE Conference on Computer Vision and Pattern Recognition}, 2019, pp. 1349--1358.

\bibitem{chib2024ms}
P.~S. Chib, A.~Nath, P.~Kabra, I.~Gupta, and P.~Singh, ``Ms-tip: Imputation aware pedestrian trajectory prediction,'' in \emph{International Conference on Machine Learning}.\hskip 1em plus 0.5em minus 0.4em\relax PMLR, 2024, pp. 8389--8402.

\bibitem{mao2023leapfrog}
W.~Mao, C.~Xu, Q.~Zhu, S.~Chen, and Y.~Wang, ``Leapfrog diffusion model for stochastic trajectory prediction,'' in \emph{Proceedings of the IEEE/CVF Conference on Computer Vision and Pattern Recognition}, 2023, pp. 5517--5526.

\bibitem{kim2024higher}
S.~Kim, H.-g. Chi, H.~Lim, K.~Ramani, J.~Kim, and S.~Kim, ``Higher-order relational reasoning for pedestrian trajectory prediction,'' in \emph{Proceedings of the IEEE/CVF Conference on Computer Vision and Pattern Recognition}, 2024, pp. 15\,251--15\,260.

\bibitem{wong2024resonance}
C.~Wong, Z.~Zou, and B.~Xia, ``Resonance: Learning to predict social-aware pedestrian trajectories as co-vibrations,'' in \emph{Proceedings of the IEEE/CVF International Conference on Computer Vision}, 2025, pp. 25\,788--25\,799.

\bibitem{wong2025reverberation}
C.~Wong, Z.~Zou, B.~Xia, and X.~You, ``Reverberation: Learning the latencies before forecasting trajectories,'' \emph{arXiv preprint arXiv:2511.11164}, 2025.

\bibitem{vaswani2017attention}
A.~Vaswani, N.~Shazeer, N.~Parmar, J.~Uszkoreit, L.~Jones, A.~N. Gomez, {\L}.~Kaiser, and I.~Polosukhin, ``Attention is all you need,'' in \emph{Advances in neural information processing systems}, 2017, pp. 5998--6008.

\bibitem{pellegrini2009youll}
S.~Pellegrini, A.~Ess, K.~Schindler, and L.~Van~Gool, ``You'll never walk alone: Modeling social behavior for multi-target tracking,'' in \emph{2009 IEEE 12th International Conference on Computer Vision}.\hskip 1em plus 0.5em minus 0.4em\relax IEEE, 2009, pp. 261--268.

\bibitem{lerner2007crowds}
A.~Lerner, Y.~Chrysanthou, and D.~Lischinski, ``Crowds by example,'' \emph{Computer Graphics Forum}, vol.~26, no.~3, pp. 655--664, 2007.

\bibitem{robicquet2016learning}
A.~Robicquet, A.~Sadeghian, A.~Alahi, and S.~Savarese, ``Learning social etiquette: Human trajectory understanding in crowded scenes,'' in \emph{European conference on computer vision}.\hskip 1em plus 0.5em minus 0.4em\relax Springer, 2016, pp. 549--565.

\bibitem{liang2020simaug}
J.~Liang, L.~Jiang, and A.~Hauptmann, ``Simaug: Learning robust representations from simulation for trajectory prediction,'' in \emph{Proceedings of the European conference on computer vision (ECCV)}, August 2020.

\bibitem{liang2020garden}
J.~Liang, L.~Jiang, K.~Murphy, T.~Yu, and A.~Hauptmann, ``The garden of forking paths: Towards multi-future trajectory prediction,'' in \emph{Proceedings of the IEEE/CVF Conference on Computer Vision and Pattern Recognition}, 2020, pp. 10\,508--10\,518.

\bibitem{caesar2019nuscenes}
H.~Caesar, V.~Bankiti, A.~H. Lang, S.~Vora, V.~E. Liong, Q.~Xu, A.~Krishnan, Y.~Pan, G.~Baldan, and O.~Beijbom, ``nuscenes: A multimodal dataset for autonomous driving,'' \emph{arXiv preprint arXiv:1903.11027}, 2019.

\bibitem{saadatnejad2022pedestrian}
S.~Saadatnejad, Y.~Z. Ju, and A.~Alahi, ``Pedestrian 3d bounding box prediction,'' \emph{arXiv preprint arXiv:2206.14195}, 2022.

\bibitem{yuan2021agentformer}
Y.~Yuan, X.~Weng, Y.~Ou, and K.~M. Kitani, ``Agentformer: Agent-aware transformers for socio-temporal multi-agent forecasting,'' in \emph{Proceedings of the IEEE/CVF International Conference on Computer Vision (ICCV)}, 2021, pp. 9813--9823.

\bibitem{linou2016nba}
K.~Linou, D.~Linou, and M.~de~Boer, ``Nba player movements,'' https://github.com/linouk23/NBA-Player-Movements, 2016.

\bibitem{xu2022remember}
C.~Xu, W.~Mao, W.~Zhang, and S.~Chen, ``Remember intentions: Retrospective-memory-based trajectory prediction,'' in \emph{Proceedings of the IEEE/CVF Conference on Computer Vision and Pattern Recognition (CVPR)}, June 2022, pp. 6488--6497.

\bibitem{yu2020spatio}
C.~Yu, X.~Ma, J.~Ren, H.~Zhao, and S.~Yi, ``Spatio-temporal graph transformer networks for pedestrian trajectory prediction,'' in \emph{European Conference on Computer Vision}.\hskip 1em plus 0.5em minus 0.4em\relax Springer, 2020, pp. 507--523.

\bibitem{mangalam2020not}
K.~Mangalam, H.~Girase, S.~Agarwal, K.-H. Lee, E.~Adeli, J.~Malik, and A.~Gaidon, ``It is not the journey but the destination: Endpoint conditioned trajectory prediction,'' in \emph{European Conference on Computer Vision}, 2020, pp. 759--776.

\bibitem{hu2020collaborative}
Y.~Hu, S.~Chen, Y.~Zhang, and X.~Gu, ``Collaborative motion prediction via neural motion message passing,'' in \emph{Proceedings of the IEEE/CVF Conference on Computer Vision and Pattern Recognition}, 2020, pp. 6319--6328.

\bibitem{salzmann2020trajectron}
T.~Salzmann, B.~Ivanovic, P.~Chakravarty, and M.~Pavone, ``Trajectron++: Dynamically-feasible trajectory forecasting with heterogeneous data,'' in \emph{Proceedings of the European conference on computer vision (ECCV)}.\hskip 1em plus 0.5em minus 0.4em\relax Springer, 2020, pp. 683--700.

\bibitem{wong2022view}
C.~Wong, B.~Xia, Z.~Hong, Q.~Peng, W.~Yuan, Q.~Cao, Y.~Yang, and X.~You, ``View vertically: A hierarchical network for trajectory prediction via fourier spectrums,'' in \emph{European Conference on Computer Vision}.\hskip 1em plus 0.5em minus 0.4em\relax Springer, 2022, pp. 682--700.

\bibitem{wu2023multi}
Y.~Wu, L.~Wang, S.~Zhou, J.~Duan, G.~Hua, and W.~Tang, ``Multi-stream representation learning for pedestrian trajectory prediction,'' in \emph{Proceedings of the AAAI Conference on Artificial Intelligence}, vol.~37, no.~3, 2023, pp. 2875--2882.

\bibitem{choi2024dice}
Y.~Choi, R.~C. Mercurius, S.~M.~A. Shabestary, and A.~Rasouli, ``Dice: Diverse diffusion model with scoring for trajectory prediction,'' in \emph{2024 IEEE Intelligent Vehicles Symposium (IV)}.\hskip 1em plus 0.5em minus 0.4em\relax IEEE, 2024, pp. 3023--3029.

\bibitem{xu2024adapting}
Y.~Xu and Y.~Fu, ``Adapting to length shift: Flexilength network for trajectory prediction,'' in \emph{Proceedings of the IEEE/CVF Conference on Computer Vision and Pattern Recognition}, 2024, pp. 15\,226--15\,237.

\bibitem{marchetti2024smemo}
F.~Marchetti, F.~Becattini, L.~Seidenari, and A.~Del~Bimbo, ``Smemo: social memory for trajectory forecasting,'' \emph{IEEE Transactions on Pattern Analysis and Machine Intelligence}, 2024.

\bibitem{lin2024progressive}
X.~Lin, T.~Liang, J.~Lai, and J.-F. Hu, ``Progressive pretext task learning for human trajectory prediction,'' in \emph{European Conference on Computer Vision}.\hskip 1em plus 0.5em minus 0.4em\relax Springer, 2024, pp. 197--214.

\bibitem{su2024unified}
Y.~Su, Y.~Li, W.~Wang, J.~Zhou, and X.~Li, ``A unified environmental network for pedestrian trajectory prediction,'' in \emph{Proceedings of the AAAI Conference on Artificial Intelligence}, vol.~38, no.~5, 2024, pp. 4970--4978.

\bibitem{liu2024uncertainty}
Y.~Liu, Z.~Ye, R.~Wang, B.~Li, Q.~Z. Sheng, and L.~Yao, ``Uncertainty-aware pedestrian trajectory prediction via distributional diffusion,'' \emph{Knowledge-Based Systems}, p. 111862, 2024.

\bibitem{wong2023another}
B.~Xia, C.~Wong, D.~Xu, Q.~Peng, and X.~You, ``Another vertical view: A hierarchical network for heterogeneous trajectory prediction via spectrums,'' \emph{IEEE Transactions on Pattern Analysis and Machine Intelligence}, 2025.

\bibitem{yang2025sopermodel}
H.~Yang, Y.~Tian, C.~Tian, H.~Yu, W.~Lu, C.~Deng, and X.~Sun, ``Sopermodel: Leveraging social perception for multi-agent trajectory prediction,'' \emph{IEEE Transactions on Geoscience and Remote Sensing}, 2025.

\bibitem{chen2021human}
G.~Chen, J.~Li, J.~Lu, and J.~Zhou, ``Human trajectory prediction via counterfactual analysis,'' in \emph{Proceedings of the IEEE/CVF International Conference on Computer Vision}, 2021, pp. 9824--9833.

\end{thebibliography}

\begin{IEEEbiography}[{
    \includegraphics[width=1in,
                     height=1.25in,
                     clip,
                     keepaspectratio]{
        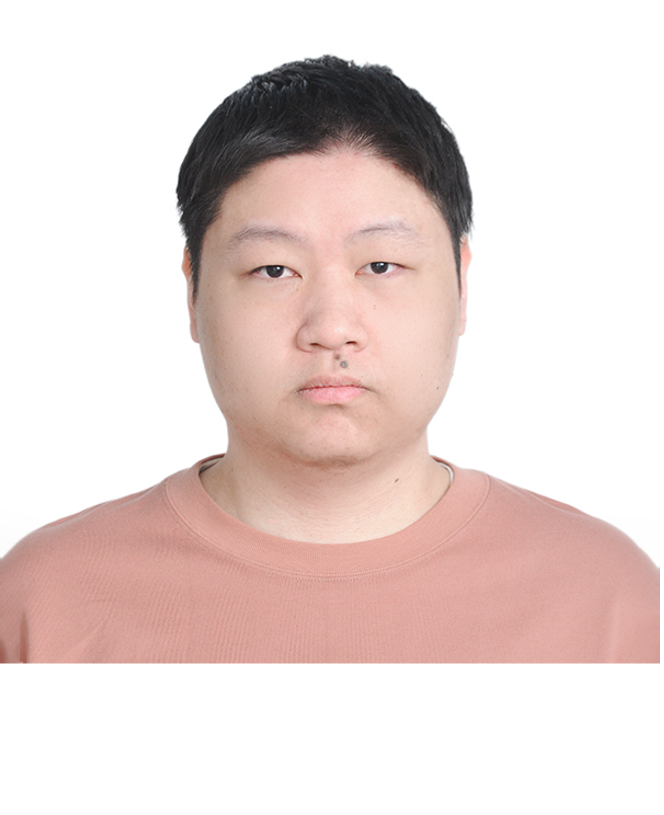
}}]{Conghao Wong}
    received the master's degree from Huazhong University of Science and Technology, Wuhan, in 2022, where he is currently pursuing the Ph.D. degree.
    His research interests include computer vision and pattern recognition.
\end{IEEEbiography}

\begin{IEEEbiography}[{
    \includegraphics[width=1in,
                     height=1.25in,
                     clip,
                     keepaspectratio]{
        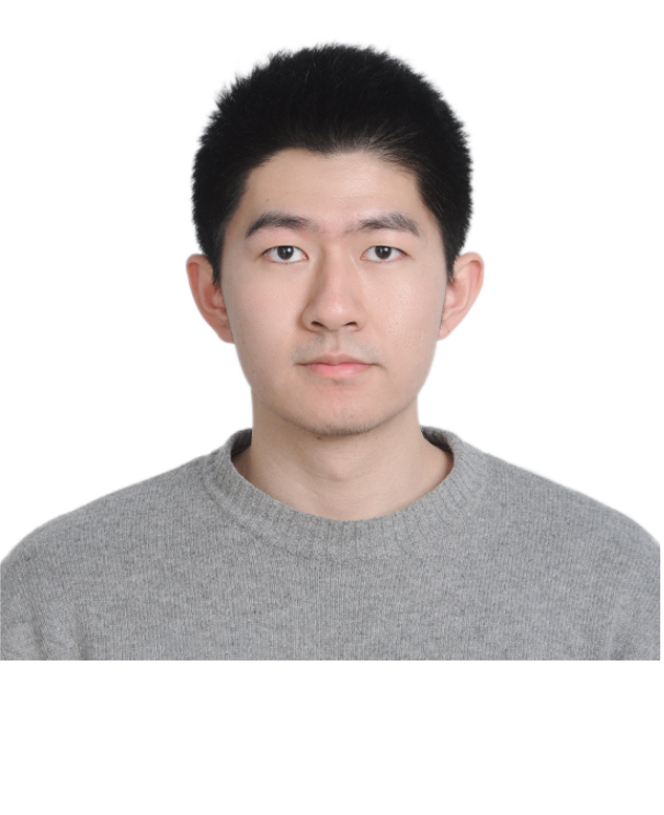
}}]{Ziqian Zou}
    received the master's degree from Huazhong University of Science and Technology, Wuhan, in 2025, where he is currently pursuing the Ph.D. degree.
    His research interests include pattern recognition and video understanding.
\end{IEEEbiography}

\begin{IEEEbiography}[{
    \includegraphics[width=1in,
                     height=1.25in,
                     clip,
                     keepaspectratio]{
        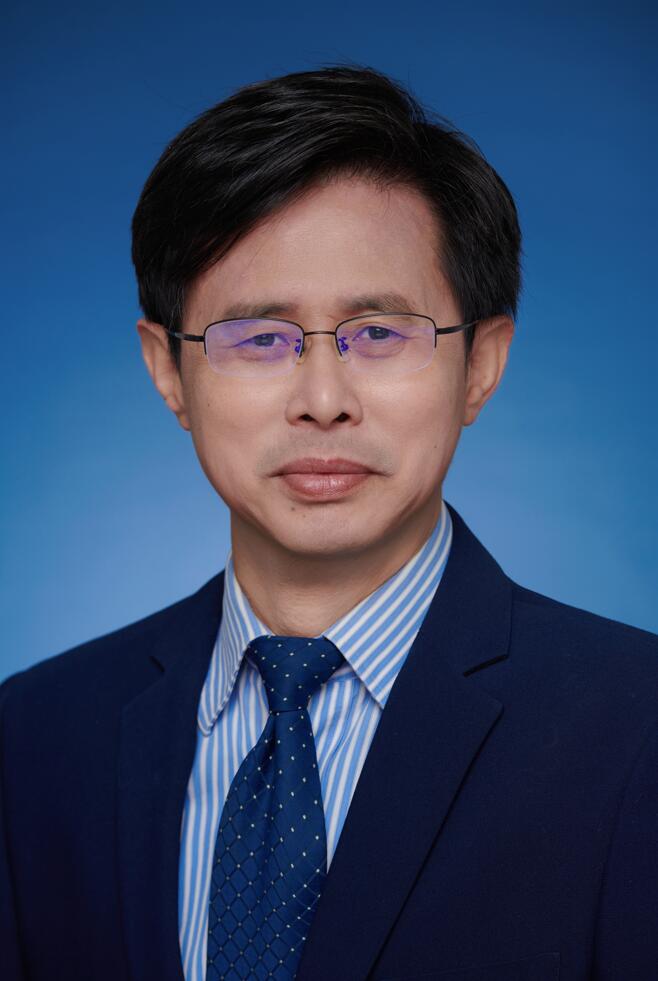
}}]{Xinge You}
    (Senior Member, IEEE) is currently a Professor in Huazhong University of Science and Technology. 
    He received his Ph.D. degree from the Department of Computer Science, Hong Kong Baptist University in 2004. 
    His work has appeared in 200+ publications, such as IEEE TPAMI, TIP, TNNLS, CVPR, ECCV, ICCV. 
    He served/serves as an Associate Editor of the \textit{IEEE TCyb}, \textit{TSMCA}. 
    His research interests include wavelet analysis, pattern recognition, machine learning, and computer vision. 
\end{IEEEbiography}

\vfill

\end{document}